\documentclass[preprint]{elsarticle}

\usepackage[utf8]{inputenc}
\usepackage{amsmath}
\usepackage{graphicx}
\usepackage{subcaption}
\usepackage{float}

\usepackage[a4paper, total={6in, 8in}]{geometry}

\pdfoutput=1

\title{Forecasting new diseases in low-data settings using transfer learning}

\author[1]{Kirstin Roster \corref{cor1}}
\ead{kirstin.roster@usp.br}
\author[2,3]{Colm Connaughton}
\author[1]{Francisco A. Rodrigues}

\address[1]{Institute of Mathematics and Computer Science, University of São Paulo, Avenida Trabalhador São Carlense 400,  São  Carlos  13566-590,  São  Paulo,  Brazil}
\cortext[cor1]{Corresponding author: kirstin.roster@usp.br}

\address[2]{Mathematics  Institute,  University  of  Warwick,  Coventry, CV4 7AL, United Kingdom}

\address[3]{London Mathematical Laboratory, 8 Margravine Gardens, W6 8RH, London,United Kingdom}

\date{}
\begin{document}

\begin{abstract}
    Recent infectious disease outbreaks, such as the COVID-19 pandemic and the Zika epidemic in Brazil, have demonstrated both the importance and difficulty of accurately forecasting novel infectious diseases. When new diseases first emerge, we have little knowledge of the transmission process, the level and duration of immunity to reinfection, or other parameters required to build realistic epidemiological models. Time series forecasts and machine learning, while less reliant on assumptions about the disease, require large amounts of data that are also not available in early stages of an outbreak. In this study, we examine how knowledge of related diseases can help make predictions of new diseases in data-scarce environments using transfer learning. We implement both an empirical and a theoretical approach. Using empirical data from Brazil, we compare how well different machine learning models transfer knowledge between two different disease pairs: (i) dengue and Zika, and (ii) influenza and COVID-19. In the theoretical analysis, we generate data using different transmission and recovery rates with an SIR compartmental model, and then compare the effectiveness of different transfer learning methods. We find that transfer learning offers the potential to improve predictions, even beyond a model based on data from the target disease, though the appropriate source disease must be chosen carefully. While imperfect, these models offer an additional input for decision makers during pandemic response.
\end{abstract}

\maketitle

\section{Introduction}

Epidemic models can be divided into two broad categories: data-driven models aim to fit an epidemic curve to past data in order to make predictions about the future; mechanistic models simulate scenarios based on different underlying assumptions, such as varying contact rates or vaccine effectiveness. Both model types aid in the public health response: forecasts serve as an early warning system of an outbreak in the near future, while mechanistic models help us better understand the causes of spread and potential remedial interventions to prevent further infections \cite{holmdahl2020wrong, kandula2018evaluation}. 
Many different data-driven and mechanistic models were proposed during the early stages of the COVID-19 pandemic and informed decision-making with varying levels of success \cite{holmdahl2020wrong, saltelli2020five, biggs2021revisiting}. This range of predictive performance underscores both the difficulty and importance of epidemic forecasting, especially early on in an outbreak.   
Yet the COVID-19 pandemic also led to unprecedented levels of data-sharing and collaboration across disciplines, so that several novel approaches to epidemic forecasting have been developed, including models that incorporate machine learning and real-time big data data streams \cite{bullock2020mapping}. Continuing on the recent trajectory of infectious disease outbreaks, such as the Zika virus in Brazil in 2015, the Ebola virus in West Africa in 2014-16, the Middle East respiratory syndrome (MERS) in 2012, and the coronavirus associated with severe acute respiratory syndrome (SARS-CoV) in 2003, suggests that further improvements to epidemic forecasting will be important for global public health. Decision-makers should be equipped with a diverse portfolio of surveillance systems to help answer a range of questions about the novel disease, including the likely trajectory of infections. Exploring the value of new methodologies can help broaden the modeler's toolkit to prepare for the next outbreak \cite{desai2019realtime, Lipsitch2019EnhancingSA}. In this study, we consider the role of transfer learning for pandemic response.

Transfer learning refers to a collection of techniques that apply knowledge from one prediction problem to solve another, generally implemented using machine learning models and with many recent applications in domains such as computer vision and natural language processing \cite{pan2010survey, tsung2018statistical}. Transfer learning leverages a model trained to execute a particular task in a particular domain, in order to perform a different task or extrapolate to a different domain. This allows the model to learn the new task with much less data than would normally be required, and is especially well-suited to data-scarce prediction problems. The underlying idea is that skills developed in one task, for example the features that are relevant to recognize human faces in images, may be useful in other situations, such as classification of emotions from facial expressions. Similarly, there may be shared features in the pattern of observed cases among similar diseases like different arboviruses or seasonal respiratory infections.

The value of transfer learning for the study of infectious diseases is relatively under-explored. The majority of existing studies on diseases remain in the domain of computer vision and leverage pre-trained neural networks to make diagnoses from medical images, such as retinal diseases \cite{roy2020transfer}, dental diseases \cite{prajapati2017classification}, or COVID-19 \cite{sufian2020survey, altaf2021novel}. Coelho and colleagues (2020) \cite{coelho2020transfer-preprint} explore the potential of transfer learning for disease forecasts. They train a Long Short-Term Memory (LSTM) neural network on dengue fever time series and make forecasts directly for two other mosquito-borne diseases, Zika and Chikungunya, in two Brazilian cities, Fortaleza and Rio de Janeiro. Even without any data on the two target diseases, their model achieves high prediction accuracy four weeks ahead. 
Gautam (2021) \cite{gautam2021transfer} uses COVID-19 data from Italy and the USA to build an LSTM transfer model that predicts COVID-19 cases in countries that experienced a later pandemic onset. 

These studies provide some limited empirical evidence that transfer learning may be a valuable tool for epidemic forecasting in low-data situations, though more research is needed. In this study, we aim to contribute to this empirical literature by comparing not only different types of knowledge transfer and forecasting algorithms, but also considering two different pairs of endemic and novel diseases observed in Brazilian cities, specifically (i) dengue and Zika, and (ii) influenza and COVID-19. With an additional analysis on simulated time series, we hope to provide theoretical guidance on the selection of appropriate disease pairs, by better understanding how different characteristics of the source and target diseases affect the viability of transfer learning. 

Zika and COVID-19 are two recent examples of novel emerging diseases. 
Brazil experienced a Zika epidemic in 2015-16 and the WHO declared a public health emergency of global concern in February 2016 \cite{who2016who}. 
Zika is caused by an arbovirus spread primarily by mosquito bites, though other transmission methods, including congenital, sexual, and via blood transfusions have also been observed. Zika belongs to the family of viral hemorrhagic fevers and symptoms of infection share some commonalities with other mosquito-borne \textit{flaviviridae}, such as yellow fever, dengue fever, or chikungunya. Illness tends to be mild but can lead to complications as well as microcephaly and other brain defects in the case of congenital transmission \cite{cdc2021flaviviridae, cdc2019zika}.

Given the similarity of the pathogen and primary transmission route, dengue fever is an appropriate choice of source disease for Zika forecasting. Not only does the shared mosquito vector result in similar seasonal patterns of annual outbreaks, but consistent, geographically and temporally granular data on dengue cases is available publicly via the open data initiative of the Brazilian government \cite{nunes2019thirty}. 

COVID-19 is an acute respiratory infection caused by the novel coronavirus SARS-CoV-2, which was first detected in Wuhan, China, in 2019. The disease has spread quickly across the globe, disrupting health systems and economies \cite{mckibbin2020economic}. It is transmitted directly between humans via airborne respiratory droplets and particles. Symptoms range from mild to fatal, and largely affect the respiratory tract. Several variants of the virus have emerged, which differ in both their severity and their transmissibility \cite{wu2020sarscov2, epaIndoorAir, cdc2021variants}.

Influenza is also a contagious respiratory disease that is spread primarily via respiratory droplets. Infection with the influenza virus also follows patterns of human contact and seasonality. There are two types of influenza (A and B) and new strains of each type emerge regularly. Given the similarity in transmission routes and to a lesser extent in clinical manifestations, influenza is chosen as the source disease for knowledge transfer to model COVID-19 \cite{cdc2021flu, hopkins2021flu}.

For each of these disease pairs, we collect time series data from Brazilian cities. Data on the target disease from half the cities is retained for testing. To ensure comparability, the test set is the same for all models. Using this empirical data, as well as the simulated time series data, we implement the following transfer models to make predictions.

\begin{itemize}
    \item \textit{Random forest:} First, we implement a random forest model which was recently found to capture well the time series characteristics of dengue in Brazil \cite{roster2021predicting}. We use this model to make predictions for Zika without re-training. We also train a random forest model on influenza data to make predictions for COVID-19. This is a direct transfer method, meaning we train the models only on data from the source disease (i.e. dengue and influenza).
    \item \textit{Random forest with TrAdaBoost:} We then incorporate the data from the target disease (i.e. Zika and COVID-19) using the TrAdaBoost algorithm together with the random forest model. This is an instance-based transfer learning method, which selects relevant examples from the source disease to improve predictions on the target disease.
    \item \textit{Neural network:} The second machine learning algorithm we deploy is a feed-forward neural network, which is first trained on data of the endemic disease (dengue / influenza) and applied directly to forecast the new disease (Zika / COVID-19).
    \item \textit{Neural network with re-training and fine-tuning:} We then retrain only the last layer of the neural network using data from the new disease (Zika / COVID-19) and make predictions on the test set. Finally, we fine-tune all the layers' parameters based on the new disease data using a small learning rate and low number of epochs. These models are examples of parameter-based transfer methods, since they leverage the weights generated by the source disease model to accelerate and improve learning in the target disease model.
    \item \textit{Aspirational baseline:} We compare these transfer methods to a model trained only on the target disease (Zika / COVID-19) without any data on the source disease. Specifically, we use half the cities in the target dataset for training and the other half for testing. This gives a benchmark of the performance in a large-data scenario, which would occur after a longer period of disease surveillance.  We include this large-data model as a measure of the lowest error achievable with the chosen algorithm for the specific disease and refer to these predictions as an aspirational baseline.
\end{itemize}

For all transfer models, we compare different levels of data availability, in order to simulate the predictive performance at different time points in the outbreak. An earlier cutoff time means that less data of the target disease is available and predictions rely more on the source disease time series. We examine how the data availability affects predictions and the relative performance of different transfer models.

The remainder of this paper is organized as follows. The models are described in more technical detail in section \ref{section:materials_methods}. Section \ref{section:results} shows the results of the simulations, the empirical Zika and COVID-19 predictions, as well as a case study of the city Rio de Janeiro in Brazil. Finally, section \ref{section:discussion} discusses practical implications of the analyses.

\section{Materials and Methods} \label{section:materials_methods}

\subsection{Data}
We leverage both empirical and synthetic epidemic data. For the empirical analysis, we use official weekly case reports at the municipal level of four diseases from the open data platform of the Brazilian government \cite{datasus}. Dengue and Zika data are collected from the Notifiable Diseases Information System (\textit{Sistema de Informação de Agravos de Notificação - SINAN}) data system  for the years spanning 2014-2020 and 2016-2020, respectively. 
SINAN also reports cases of Severe Acute Respiratory Syndrome (SARS), reliably across municipalities from 2013 onwards. SARS in this context refers to individuals with symptoms such as fever or difficulty breathing, which may be caused by different pathogens, including an influenza virus or a coronavirus. The case record may be accompanied by diagnostic tests. For this study, we include only the SARS cases with laboratory-confirmed influenza (strains A and B). Daily COVID-19 case reports are collected from March 2020 to September 2021 \cite{painelcovid}. They are aggregated to weekly case counts to match the reporting frequency of the other datasets.

Synthetic time series data is generated using a stochastic epidemic model with compartments for susceptible, infectious, recovered, and dead population groups (SIRD) given by equations \ref{equ:sird}.
\begin{equation}
\begin{aligned}
    \frac{dS}{dt} &= - \frac{\beta}{N}SI + \zeta R\\
    \frac{dI}{dt} &= \frac{\beta}{N}SI - \gamma I - \mu I \\
    \frac{dR}{dt} &= \gamma I - \zeta R \\
    \frac{dD}{dt} &= \mu I
\end{aligned}
\label{equ:sird}
\end{equation}
where: $\beta$ is the effective contact rate (or transmission rate), $\gamma$ is the recovery rate, $\zeta$ is the waning immunity rate, and $\mu$ is the disease-specific death rate. Infection, recovery, waning immunity, and death from disease are treated as stochastic events. At each time step, the number of occurrences of each event are sampled from a binomial distribution with probabilities given by the SIRD model parameters (equation \ref{equ:sird}) as $e^{-\beta I/N}$, $e^{-\gamma}$, $e^{-\zeta}$, and $e^{-\mu}$, respectively.

To simulate different disease settings, we vary the model parameters. Our endemic disease has parameters $\beta = 0.191, \gamma = 0.05, \zeta = 0.008, \mu = 0.0294$, while the target diseases (for transfer) have transmission and recovery rates $\beta \in \{0.25, 0.3, 0.35\}$ and $\gamma \in \{0.01, 0.1, 0.15\}$, respectively. This results in nine target diseases, each with 100 observations over 1,000 time steps. The parameter ranges were chosen from estimates in the literature of an empirical COVID-19 model \cite{anastassopoulou2020databased}. For each set of parameters, we generate two sets of 100 iterations of length $T=1000$ one for training and another for testing. Each time step in the simulation may be interpreted as one week in reality.

In the context of this simulated data, we include in the term \textit{new disease} any scenario that may result in variation of parameters, including new virus strains or changes in the contact rates within a population. For simplicity, we refer to each parameter combination as a separate disease to mimic the empirical analysis that does use data from separate diseases. 
\begin{figure}[H]
    \centering
    \includegraphics[width=0.6\textwidth]{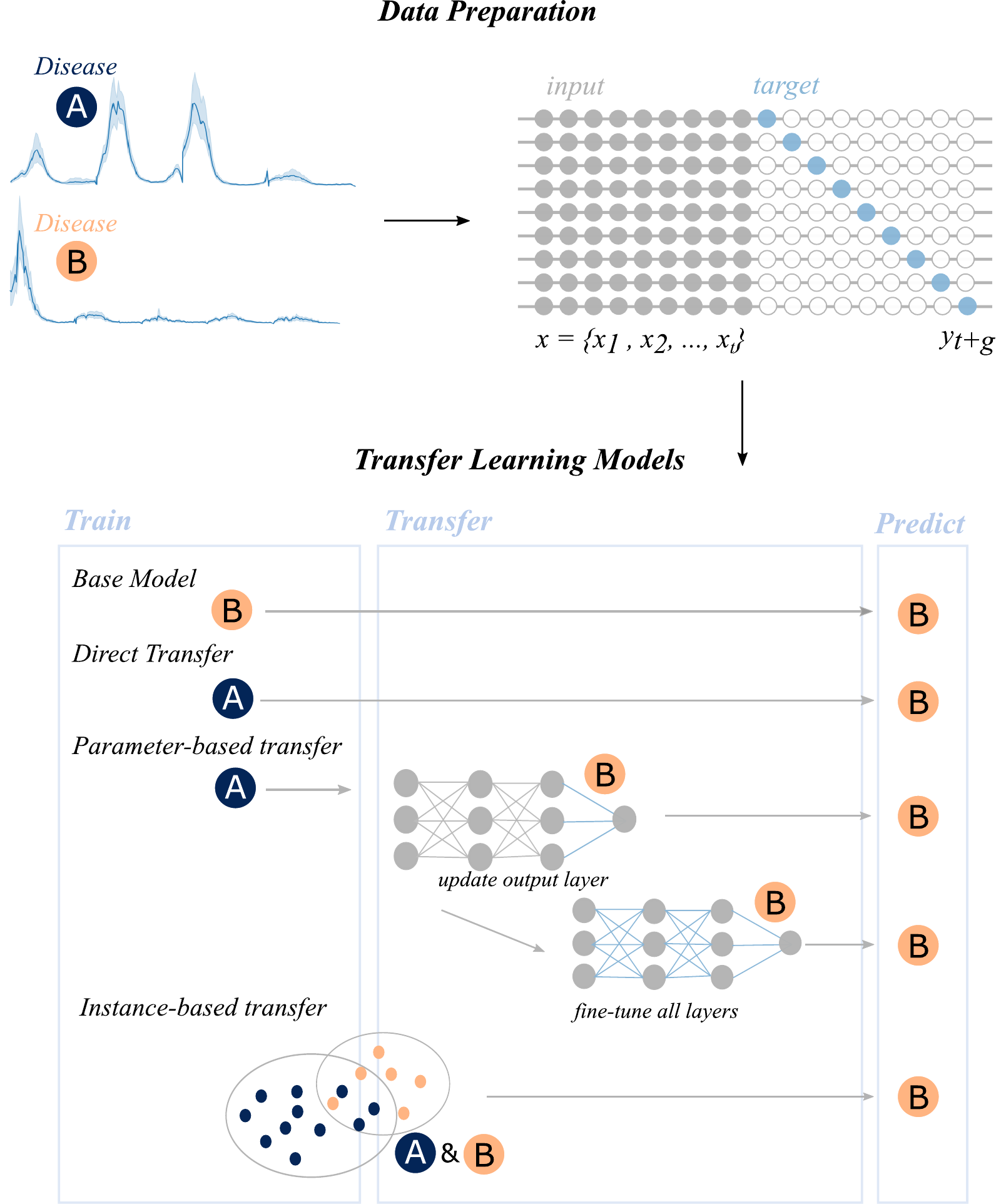}
    \caption{Methodology}
    \label{fig:methodology}
    \small
    The time series for the source (A) and target disease (B) are prepared for forecasting up to 9 time steps ahead using 9 time steps as input features. The data is used to train four kinds of models: the base model trains and tests on the target disease. Direct transfer entails training on the source disease and predicting the target disease. The parameter-based transfer model trains on the source disease and uses the target disease for parameter update in the last layer of a neural network as well as for prediction. This method may also include an additional fine-tuning step where all layers are updated. Instance-based transfer selects features from the source disease to complement the target disease data. The model is trained on the combined data and tested on the target disease.
\end{figure}

\subsection{Data Preparation}
We split both the simulated and the empirical data of the target diseases into train and test sets. In the case of simulated data, we generate separate training and test sets of equal size. In the empirical analysis, we split the cities randomly in half in order to generate equally sized train and test sets. The resulting test datasets contain 1,527 cities for Zika and 2,795 for COVID-19. The train datasets contain 217 cities for influenza and 4,587 for dengue, as well as the other half of Zika and COVID-19 cities (1,527 and 2,795, respectively).

The time series are split into short sections which are used to train the machine learning models (see figure \ref{fig:methodology}). Each section consists of the last nine observations (input features) and the future number of cases (target feature). The target feature is between two and nine time steps ahead of the input features, allowing us to compare prediction horizons up to nine weeks ahead. To ensure a fair comparison of performance across the horizons, the first eight time periods are removed from all prediction datasets. 

Transfer learning will be most useful and most challenging early on in an outbreak, when little data has been collected. We compare how the quantity of data of the new disease available for training affects performance of the transfer models, by comparing three different cutoff dates of the empirical data, four weeks apart. For the synthetic data, we compare cutoff periods of 25, 30, 35, and 100 steps of the 1000-step long time series.

\subsection{Methods}

In both the theoretical and empirical analyses, we compare three different kinds of transfer learning, eight different forecast horizons, and four different levels of data availability using the methodology presented in figure \ref{fig:methodology}. This section describes each of these models in more detail. The python code used for the analysis is available at \textit{github.com/KRoster/transfer-learning-pandemic-preparedness}.

\textbf{No transfer}. We first train a machine learning model on the endemic disease and then use it directly to forecast the target disease, without any adjustment of the model weights. Two algorithms were chosen for direct transfer: (i) a random forest algorithm with 50 trees and (ii) a fully-connected feed-forward neural network with three hidden layers and (64,32,32) neurons in the respective layers.

Random forests (RF) \cite{Breiman2001randomforests} is an ensemble method that aggregates predictions from multiple decision trees. A decsion tree determines the best splitting values of the input features in order to separate the observations according to the output value. The splitting criterion is the mean squared error (MSE). Random forests grows such trees iteratively, introducing variation by using only a subset of the observations for each tree (bagging) and a subset of the features at each branch in the tree.

Feed-forward neural networks (NN)  (\cite{haykin1999neural}) are constructed from several layers of neurons (also known as processing units or hidden units), each of which combines and transforms input features and passes them on to the next layer of neurons. Transformations entail weighting and linearly combining the vector of input features together with a bias term, and then applying a non-linear function (activation). The weights and biases are optimized to minimize the prediction errors according to a loss function, in this case the MSE, using gradient descent. 

\textbf{Transfer}. We implement one instance-based and one parameter-based transfer approach. Using the TrAdaBoost algorithm \cite{pardoe2010boosting}, the instance-based method selects observations from the source data (endemic disease) which will help improve predictions on the target data (new disease). We implement TrAdaBoost with the RF algorithm for 10 boosting iterations. At each iteration, the TrAdaBoost algorithm adjusts the importance assigned to each observation in the source dataset, so as to reduce the differences in distributions in the source and target domains. Examples of the source disease time series that are dissimilar to the target disease time series therefore receive a smaller weight and have less influence on model training. Source examples that are informative for understanding the target disease pattern receive a higher weight and are more influential in training the target model \cite{tsung2018statistical}.  

Using a neural network architecture, the parameter-based approach instead works as a warm start to training, by maintaining the parameters of the source disease model and updating only the weights in the last layer of the neural network with data from the target disease \cite{goodfellow2016deep}. We freeze the parameters in the first layers and update the output layer for 500 epochs with early stopping and exponential learning rate decay. 

\textbf{Transfer and Fine-tuning}. Finally, after updating the weights of the last layer of the neural network, we unfreeze the weights of the remaining layers and update all weights using a very low learning rate ($\alpha=0.00001$) and short training time (10 epochs). 

\textbf{Aspirational Baseline}. In order to assess performance of the transfer models, we also implement a baseline model that is both trained and tested on the target disease. This gives us an indication of the lowest possible error achievable by the machine learning algorithm for the given task. We use the same machine learning algorithms and architectures as in the direct transfer case. To train the aspirational baseline models, we use target data from half the cities in the target dataset over the full time period. We test the baseline models on the same dataset as the transfer models, consisting of the other half of cities. Given the overlapping time period used for training and testing as well as the relatively long time series, these baseline models are not representative of a model that would be available at the early stage of an outbreak. Rather, we include this model as a benchmark for the performance the given ML algorithm could have achieved in hindsight if the data on the new disease had been available.

\section{Results} \label{section:results}

\subsection{Simulations}
\begin{figure}[H]
    \centering
    \includegraphics[width=0.9\textwidth]{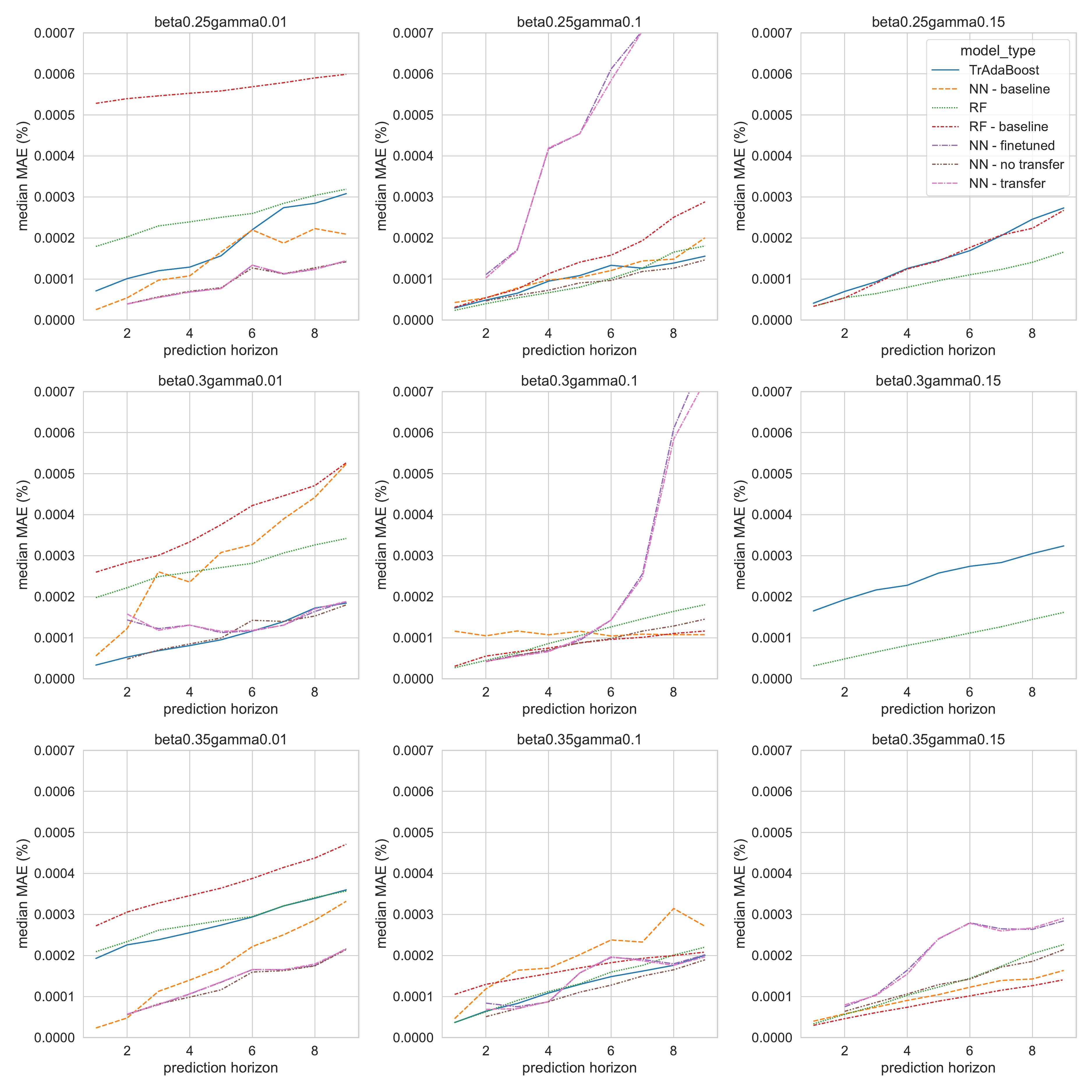}
    \caption{Mean absolute errors of predictions of synthetic datasets at data cutoff 1}
    \label{fig:synthetic_errors}
    \small
    Note: Models not shown have error values outside the y-axis range
\end{figure}

Figure \ref{fig:synthetic_errors} shows the prediction errors for the nine synthetic datasets, comparing the aspirational baseline models, the direct transfer models, as well as the transfer models at the lowest level of data availability.
The raw mean absolute error size varies largely by the different outbreak sizes across diseases, which we correct for by dividing the errors by the total number of cases in each city. We may thus interpret the metric as a percent error.

The results show that transfer learning has the potential to improve predictions even over the aspirational baseline models, which are based on the target disease data (\textit{RF - baseline} and \textit{NN - baseline}). This is especially the case for diseases whose parameters are close to the source disease. For example, the top left panel in figure \ref{fig:synthetic_errors} was generated with an effective contact rate $\beta=0.25$ (c.f. source effective contact rate $\beta=0.191$) and a recovery rate  $\gamma=0.01$ (c.f. source recovery rate $\gamma=0.05$). The \textit{NN transfer} and \textit{no transfer} models outperform both the RF and NN baselines for all prediction horizons. Conversely, when the source and target diseases differ more widely in the epidemiological parameters, the baseline models outperform the transfer models. For example, the bottom right panel in figure \ref{fig:synthetic_errors} shows that highest predictive performance is achieved by the RF baseline, followed by the NN baseline. The next best result is achieved by models trained only on the source data (\textit{RF} and \textit{NN - no transfer}).
This confirms the intuitive notion that the similarity of diseases matters to the potential value of transfer learning for disease forecasting. When the similarity cannot be known, an ensemble approach of both the direct transfer (endemic disease) and weight- or instance-based methods may be most promising.

\begin{figure}[H]
    \centering
    \includegraphics[width=0.7\textwidth]{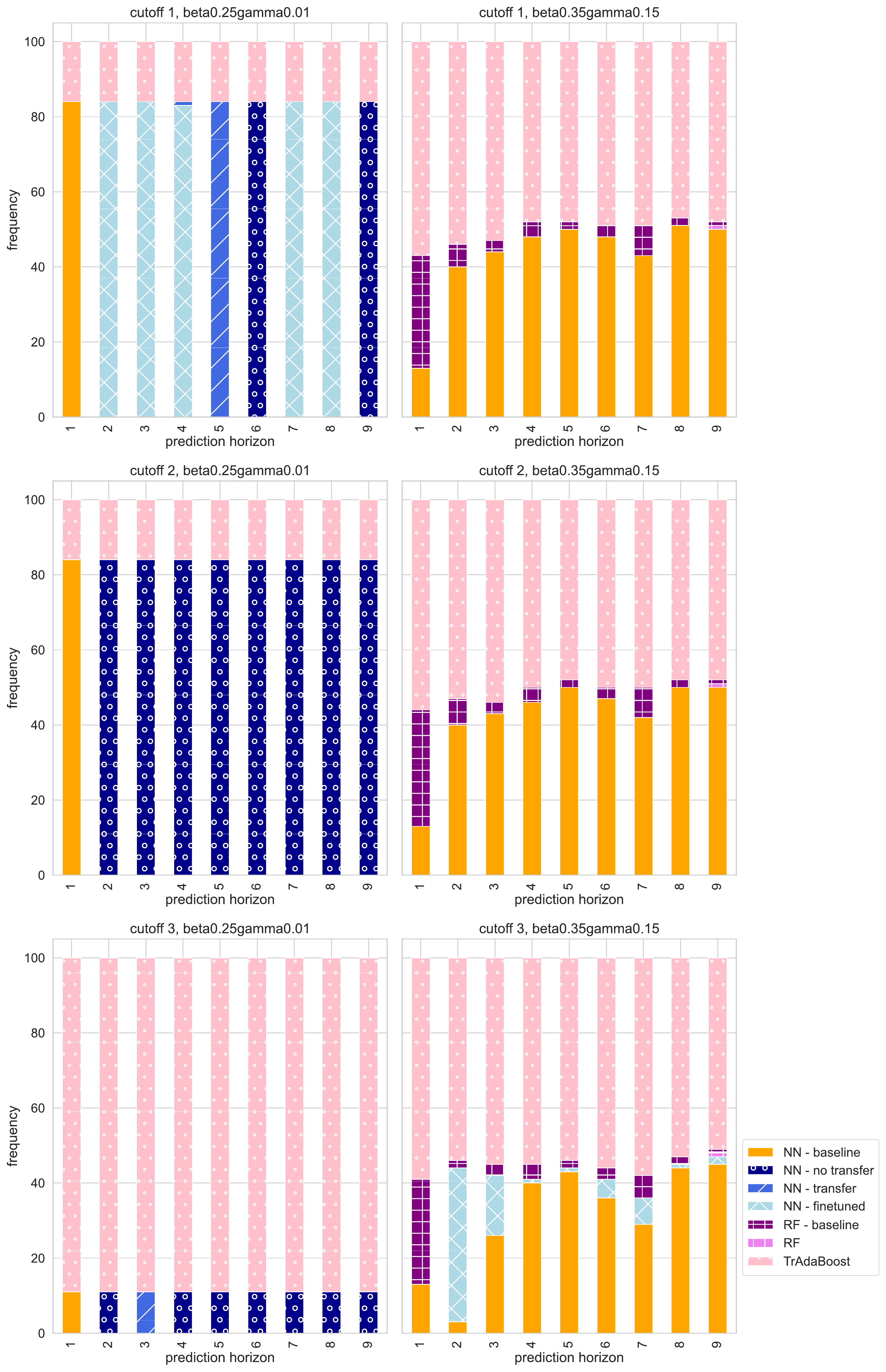}
    \caption{Frequency of best performing transfer model across prediction horizons and data cutoff levels for the target diseases most and least similar to the source disease (left and right columns, respectively)}
    \label{fig:best_model_synthetic}
\end{figure}

For each city, we identify the type of transfer model with the lowest error for different prediction horizons and levels of data availability. Figure \ref{fig:best_model_synthetic} shows how often each algorithm was chosen as the best model.
The TrAdaBoost algorithm is most popular, highlighting the benefit of allowing the algorithm to select the amount of source data useful for target disease predictions.
As data availability increases, models leveraging the target data are more frequently selected, especially in the case of the disease with less similar epidemiological parameters (right column in figure \ref{fig:best_model_synthetic}), where we observe a shift from the \textit{NN - no transfer} model to the \textit{NN - finetuned} model. In the case of the similar disease (left column), we note increased preference toward the TrAdaBoost algorithm relative to the NN models.
Overall, we note the frequent selection of the \textit{NN - no transfer} algorithm. Models leveraging none of the target data are surprisingly effective.

\begin{figure}[H]
    \centering
    \includegraphics[width=0.3\textwidth]{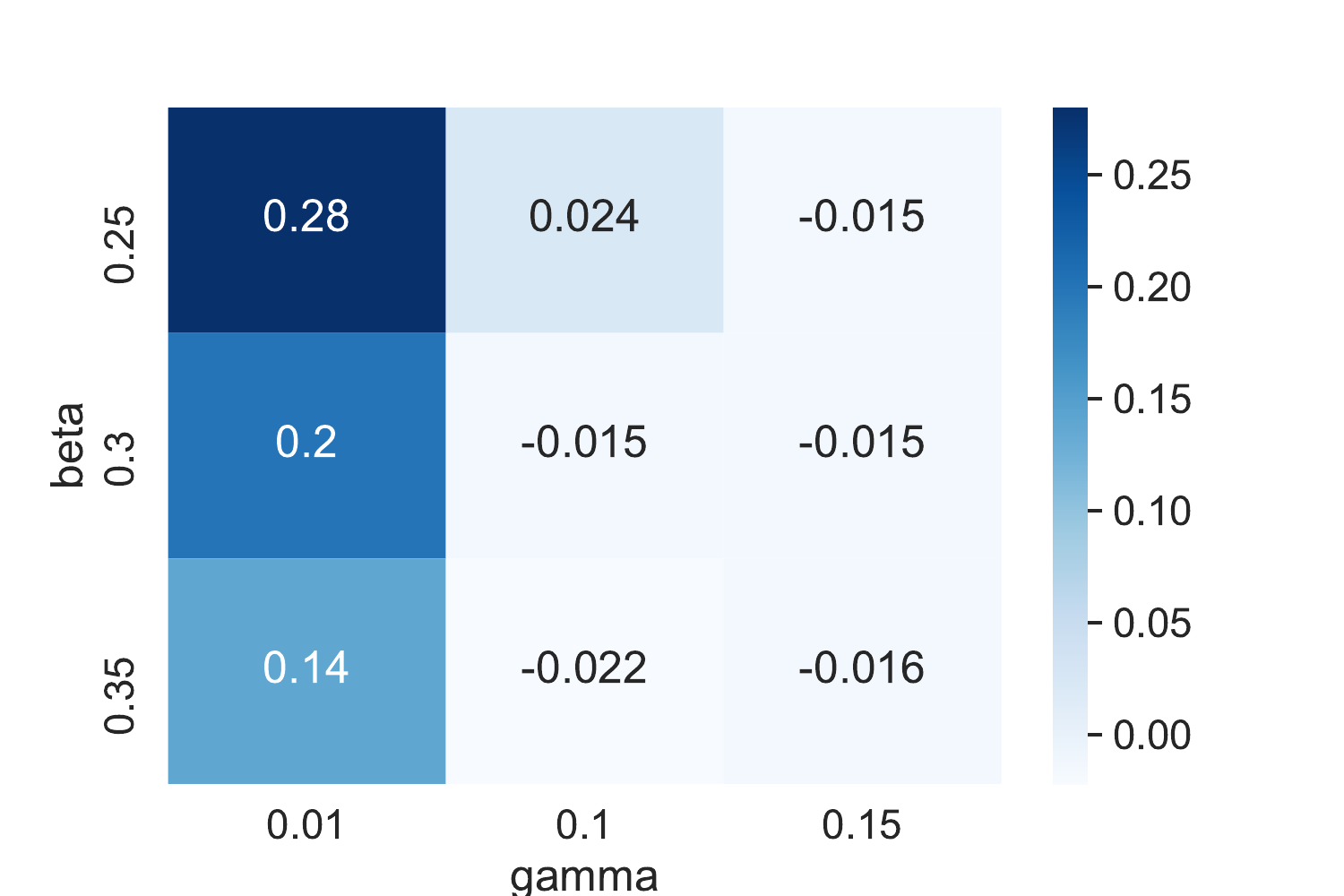}
    \caption{Median pairwise correlation coefficient of source time series with simulated datasets of varying effective contact ($\beta$) and recovery ($\gamma$) rates }
    \label{fig:corr_simulated}
\end{figure}

In lieu of known disease parameters (e.g. transmission and recovery rates), we may leverage data-based similarity measures to assess ex ante the benefit of transfer learning in general, and the different algorithms, in particular. A simple, yet effective measure is Pearson's correlation. Figure \ref{fig:corr_simulated} shows the median pairwise correlation between the time series of the source disease and each of the simulated diseases, where the x- and y-axis show the different $\gamma$- and $\beta$-values used to generate the data, respectively. The top left square represents the most similar disease in terms of the epidemiological parameters, which is also most strongly correlated with the source data. The bottom right square represents the least similar disease and has a correlation coefficient close to zero.

\subsection{Empirical Analysis}

\begin{figure}[H]
    \begin{subfigure}[t]{\textwidth}
    \centering
        \includegraphics[width=0.6\linewidth]{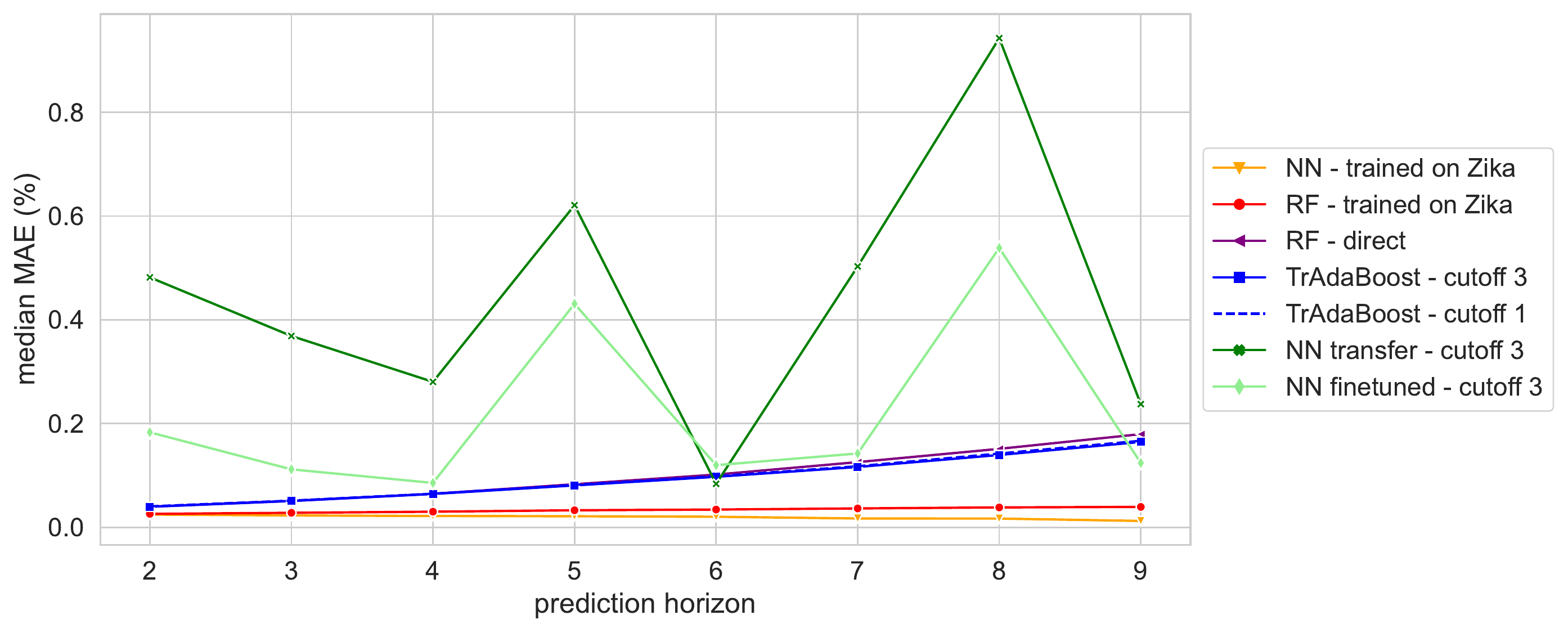}
        \caption{Zika}
    \end{subfigure}
    \begin{subfigure}[t]{\textwidth}
    \centering
        \includegraphics[width=0.6\linewidth]{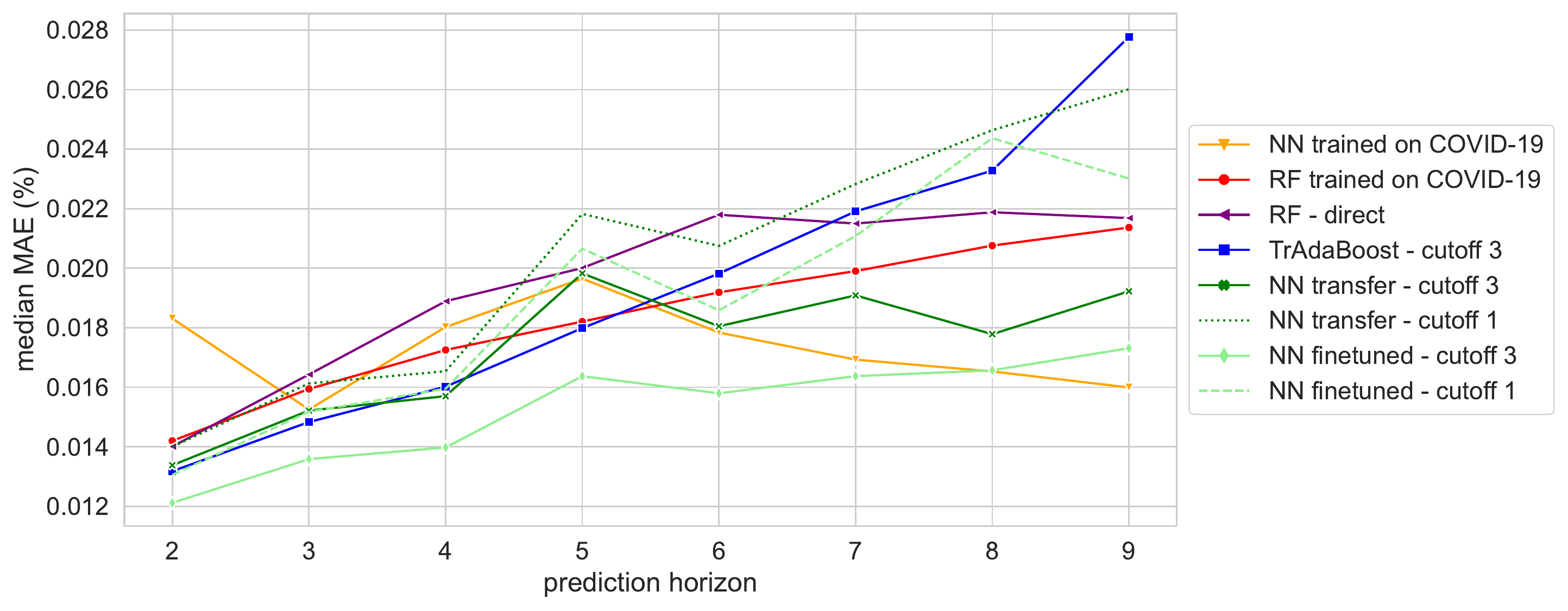}
        \caption{COVID-19}
    \end{subfigure}
    \caption{Comparison of model performance for each algorithm and disease type}
    \label{fig:select_best_model}
\end{figure}

Figure \ref{fig:select_best_model} compares the performance of the different model types for the two disease pairs, using real-world rather than simulated data. We note that for both Zika and COVID-19, models with transfer outperform the models without transfer. 

\begin{figure}[H]
    \centering
    \begin{subfigure}[t]{0.45\textwidth}
    \centering
        \includegraphics[width=0.8\linewidth]{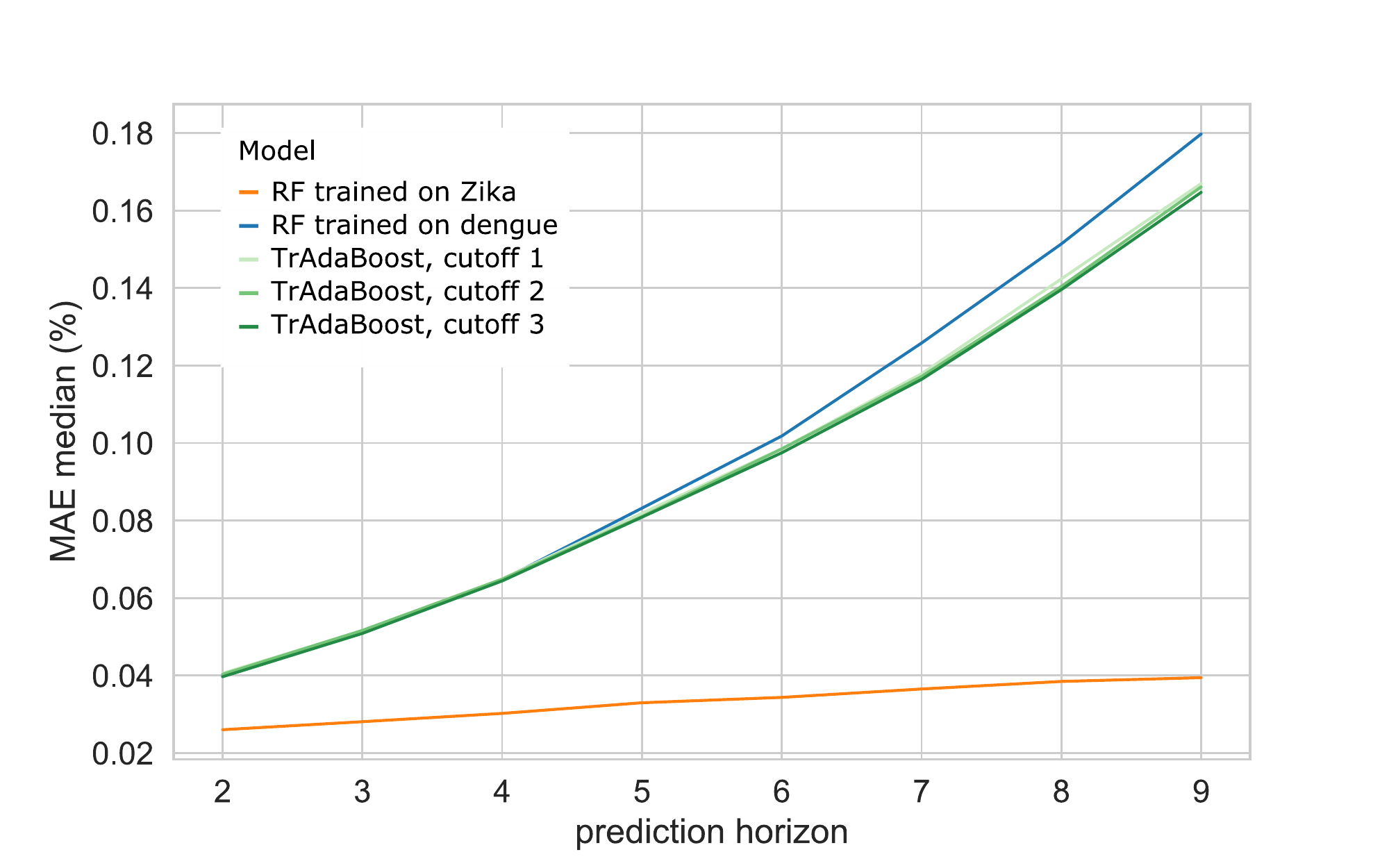}
        \caption{Zika} \label{fig:RF_errors_zika}
    \end{subfigure}
    \vspace{0.5cm}
    \begin{subfigure}[t]{\textwidth}
    \centering
        \includegraphics[width=0.8\linewidth]{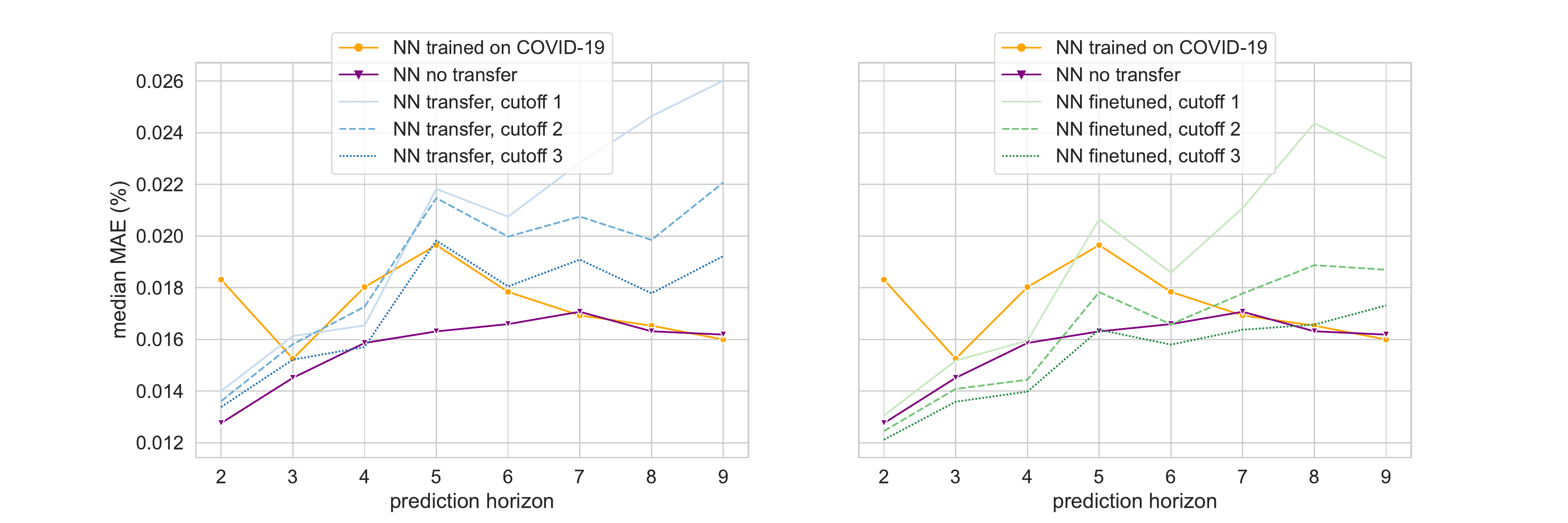}
        \caption{COVID-19} \label{fig:NN_errors_covid}
    \end{subfigure}
    \caption{Percent mean absolute errors for different prediction horizons and model types within the best-performing algorithm of each disease}
    \label{fig:cutoffs_best model_zikacovid}
    \small
    The top panel compares the different random forest models implemented for Zika prediction: TrAdaBoost with three different cutoff dates (darker shades of green indicate larger training sets), random forest trained on dengue and applied directly without transfer (blue), and the baseline random forest model trained on half of the cities of the Zika dataset (orange).\\
    The lower panel compares the different neural network models implemented for COVID-19 prediction: the left plot shows models with weight updates in the last layer of the neural network at different cutoff levels (darker shades of blue indicate larger training sets), while the right plot shows models with an extra finetuning step of all weights in addition to retraining the last layer at different data cutoff levels (darker shades of green indicate larger training sets). Both plots show the baseline model trained on half of the cities in the COVID-19 dataset (orange) and the flu-trained neural network without parameter updates (purple). 
\end{figure}

The RF models, specifically transfer with the TrAdaBoost algorithm, perform best in forecasting Zika. The finetuned transfer NN model has similarly low errors for some of the prediction horizons, but has lower performance for 5-week and 8-week predictions. The performance of the TrAdaBoost algorithm depends very little on the quantity of data available on Zika (see figure \ref{fig:RF_errors_zika}). Similarly, the RF model trained only on dengue performs surprisingly well, with nearly the same median error rate up until a prediction horizon of five weeks. For later horizons, it has only slightly decreased performance. This suggests that the dengue model can achieve nearly equal performance as the transfer model very early on in the new disease outbreak. However, at increasing prediction horizons the dengue and TrAdaBoost models increasingly deviate from the aspirational baseline model, which is trained only on Zika. 

In the case of COVID-19, the finetuned NN transfer model had the highest performance (figure \ref{fig:select_best_model}). At data cutoff level 3, the NN model with transfer and fine-tuning outperformed both aspirational baseline models - the RF and NN models trained on COVID-19 data for prediction horizons of up to 8 weeks ahead. This suggests that transfer learning may not only help approximate the performance of a model trained directly on the target disease, but may actually improve predictions due to the relatively larger overall training dataset that is available on the source disease (in this case, influenza). 

Figure \ref{fig:NN_errors_covid} shows the COVID-19 prediction errors of the NN models at different levels of data availability. Unsurprisingly, in both the weight transfer model (left) and the fine-tuned model (right), the error declines when more target disease data is available. For very early prediction horizons, all cutoff levels produce higher performance than the aspirational baseline trained on COVID-19 data (week 2 for the NN transfer model and weeks 2-4 for the NN fine-tuned model). 
As in the Zika case above, the model trained on influenza data without weight transfer has a surprisingly strong performance. It outperforms the aspirational baseline for most prediction horizons. Again, this enables predictions earlier in the outbreak, when less data on the new disease has been collected. 

\begin{figure}[H]
    \centering
    \begin{subfigure}[t]{0.45\textwidth}
        \includegraphics[width=\linewidth]{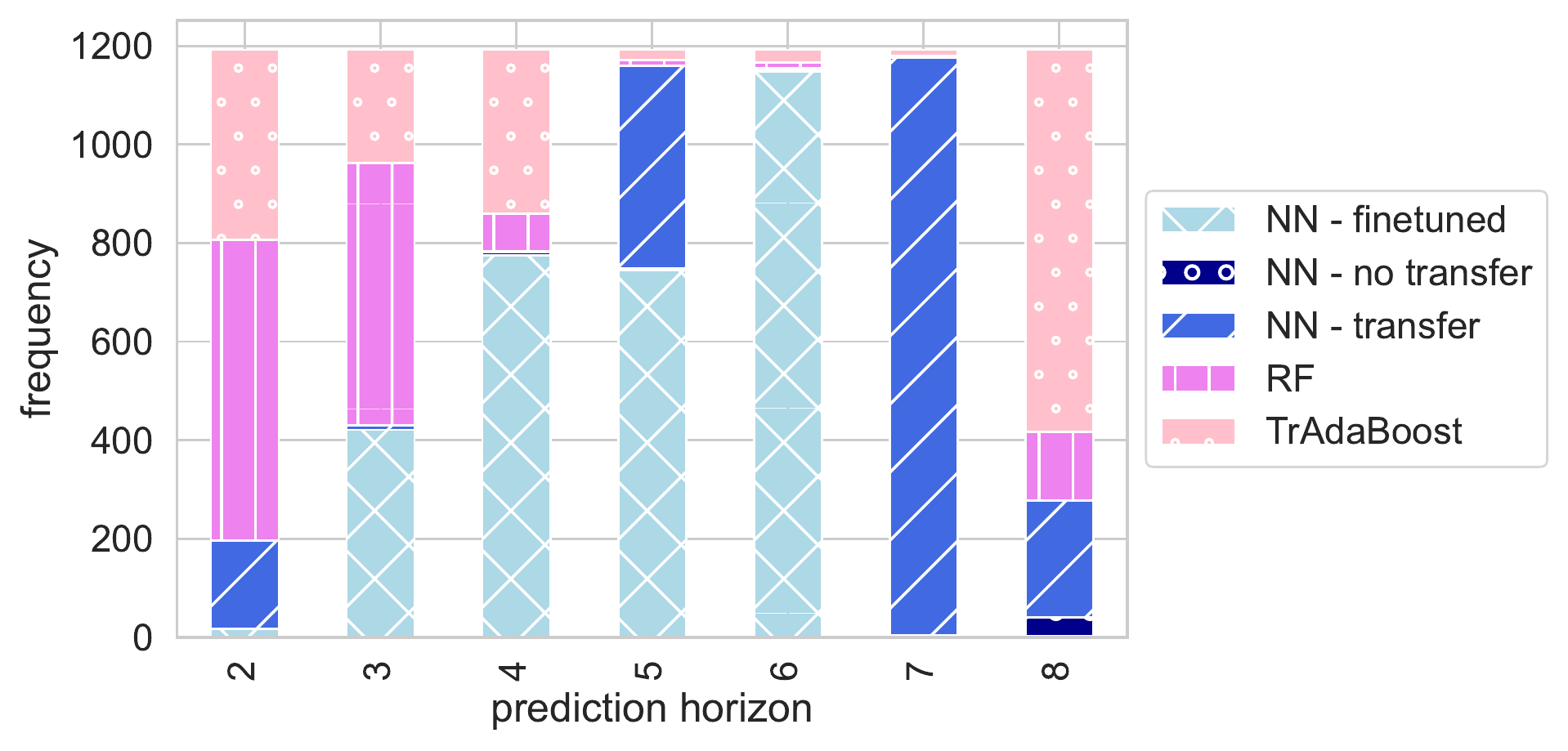}
    \caption{Zika, cutoff 1}
    \end{subfigure}
    \begin{subfigure}[t]{0.45\textwidth}
        \includegraphics[width=\linewidth]{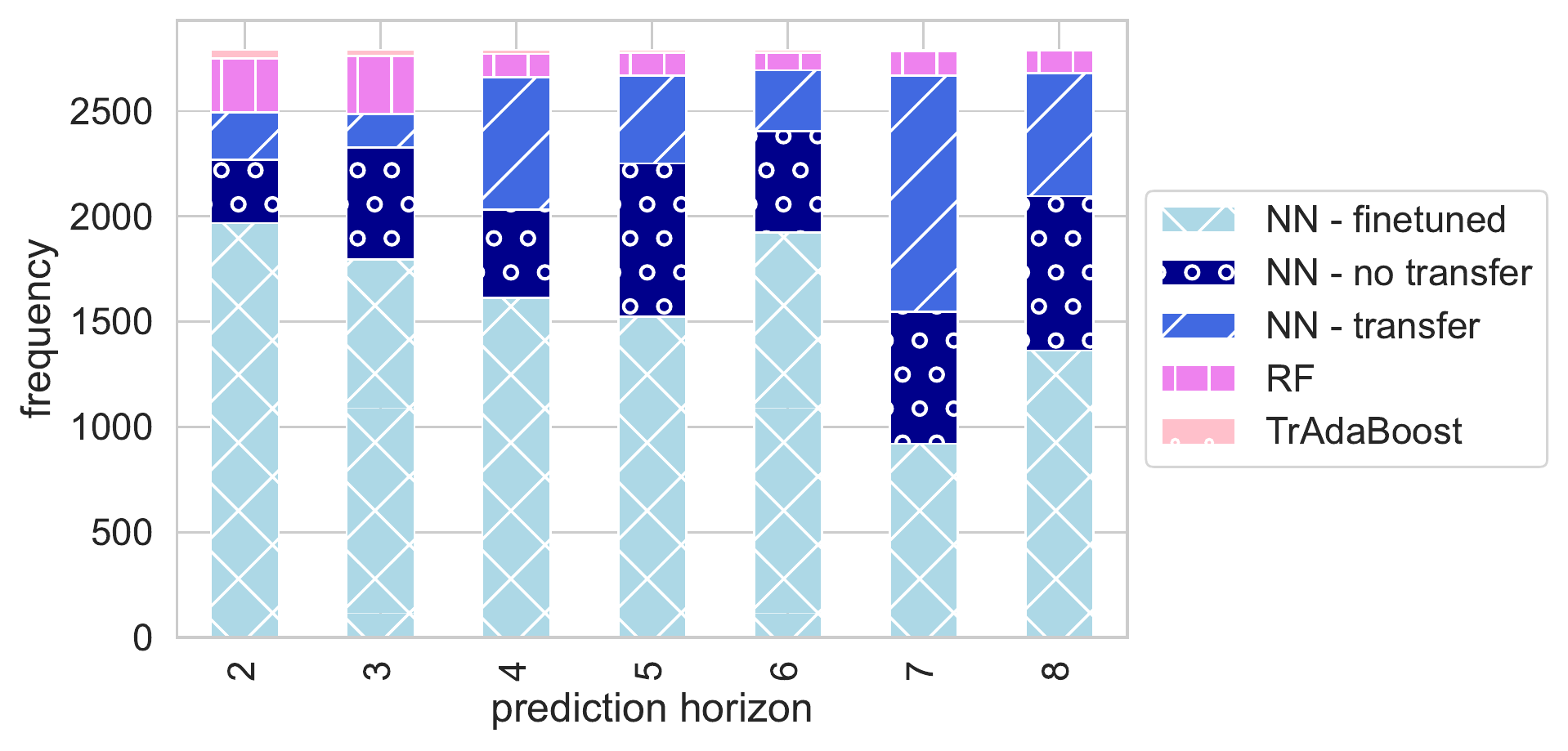}
    \caption{COVID-19, cutoff 1}
    \end{subfigure}
    \caption{Frequency of models chosen as best overall predictors of each disease at cutoff 1}
    \label{fig:my_label}
\end{figure}

\subsection{Empirical Case Study: Rio de Janeiro}

\begin{figure}[H]
    \centering
    \includegraphics[width=0.5\textwidth]{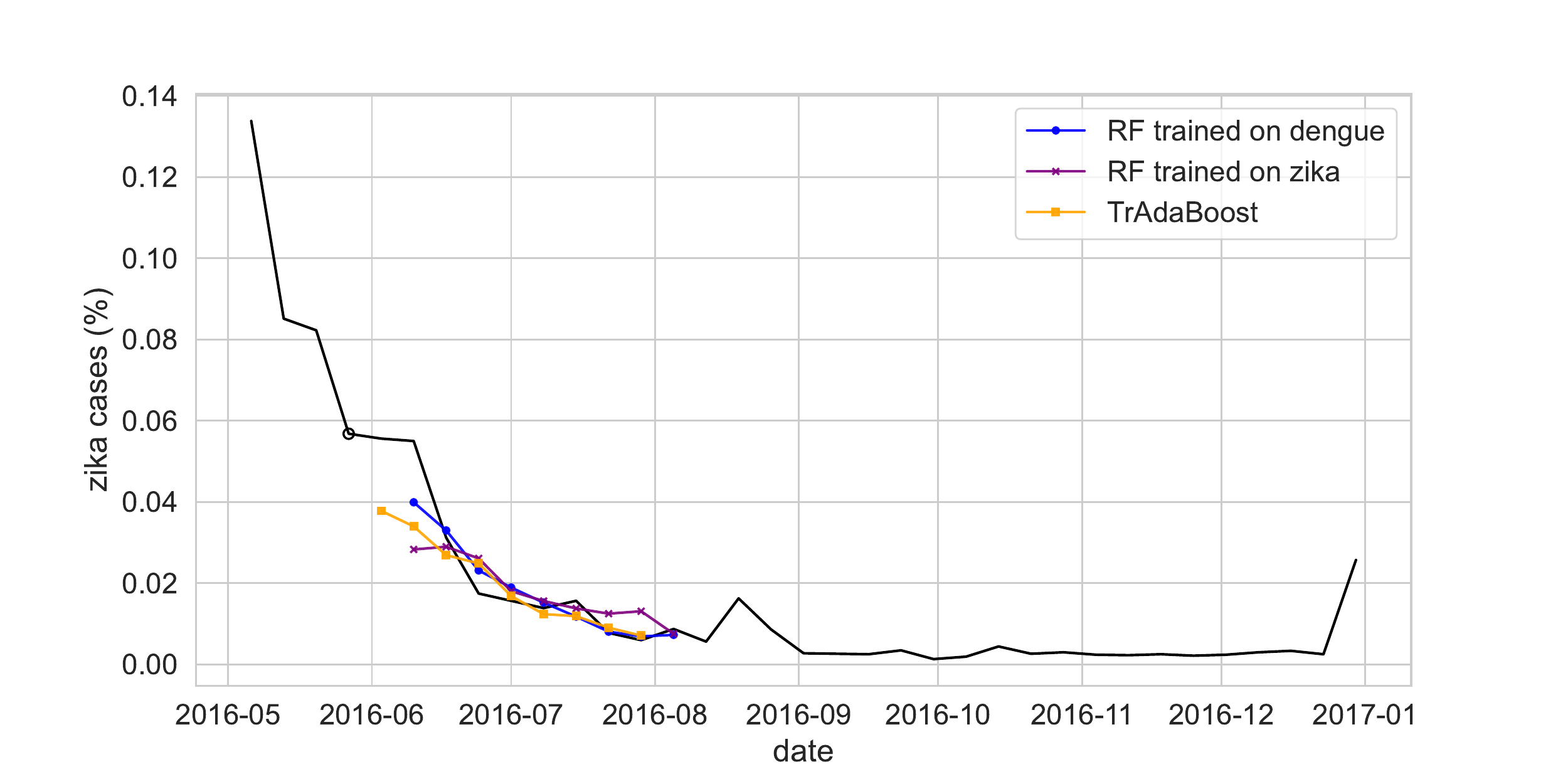}
    \includegraphics[width=0.5\textwidth]{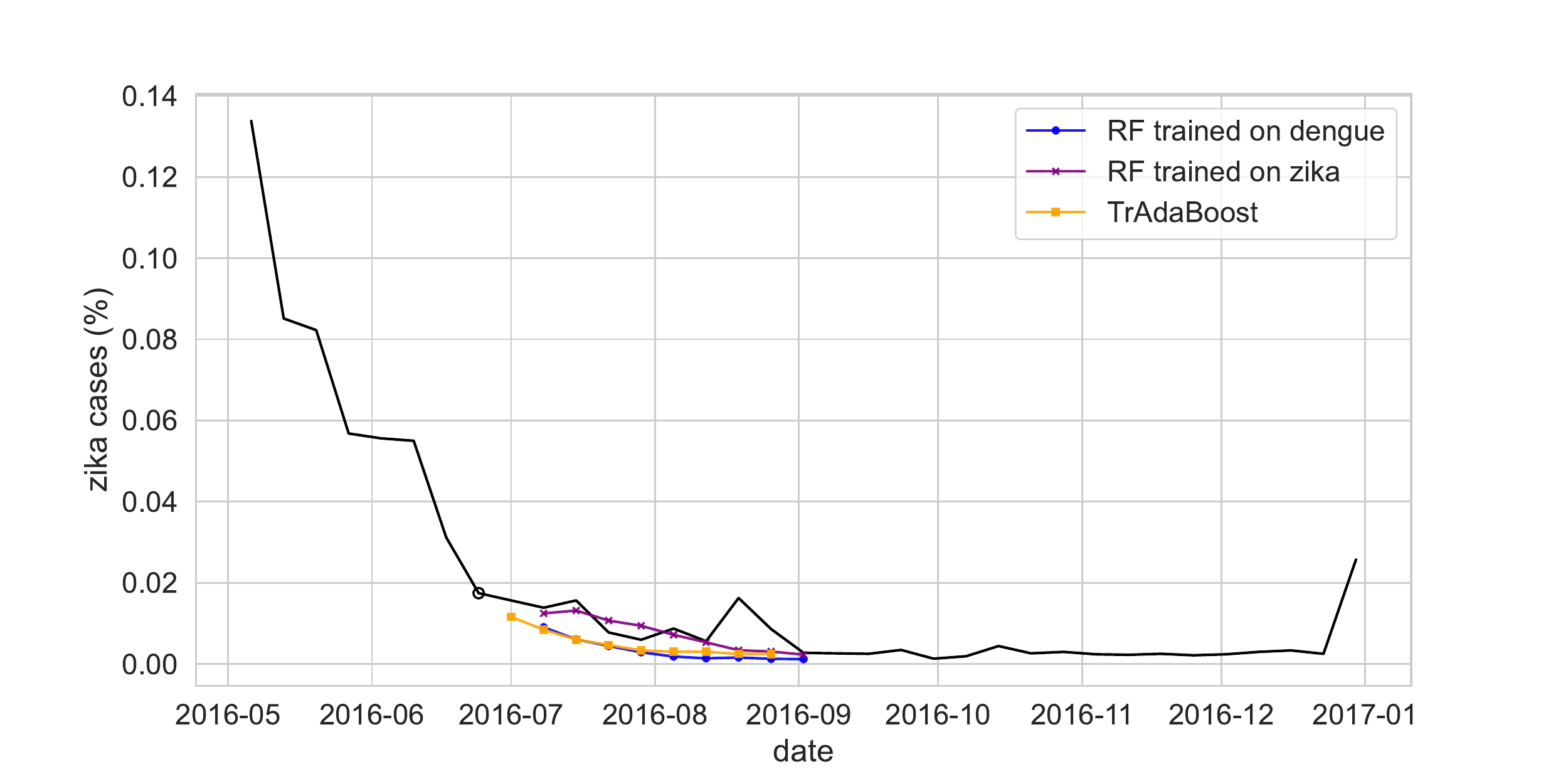}
    \includegraphics[width=0.5\textwidth]{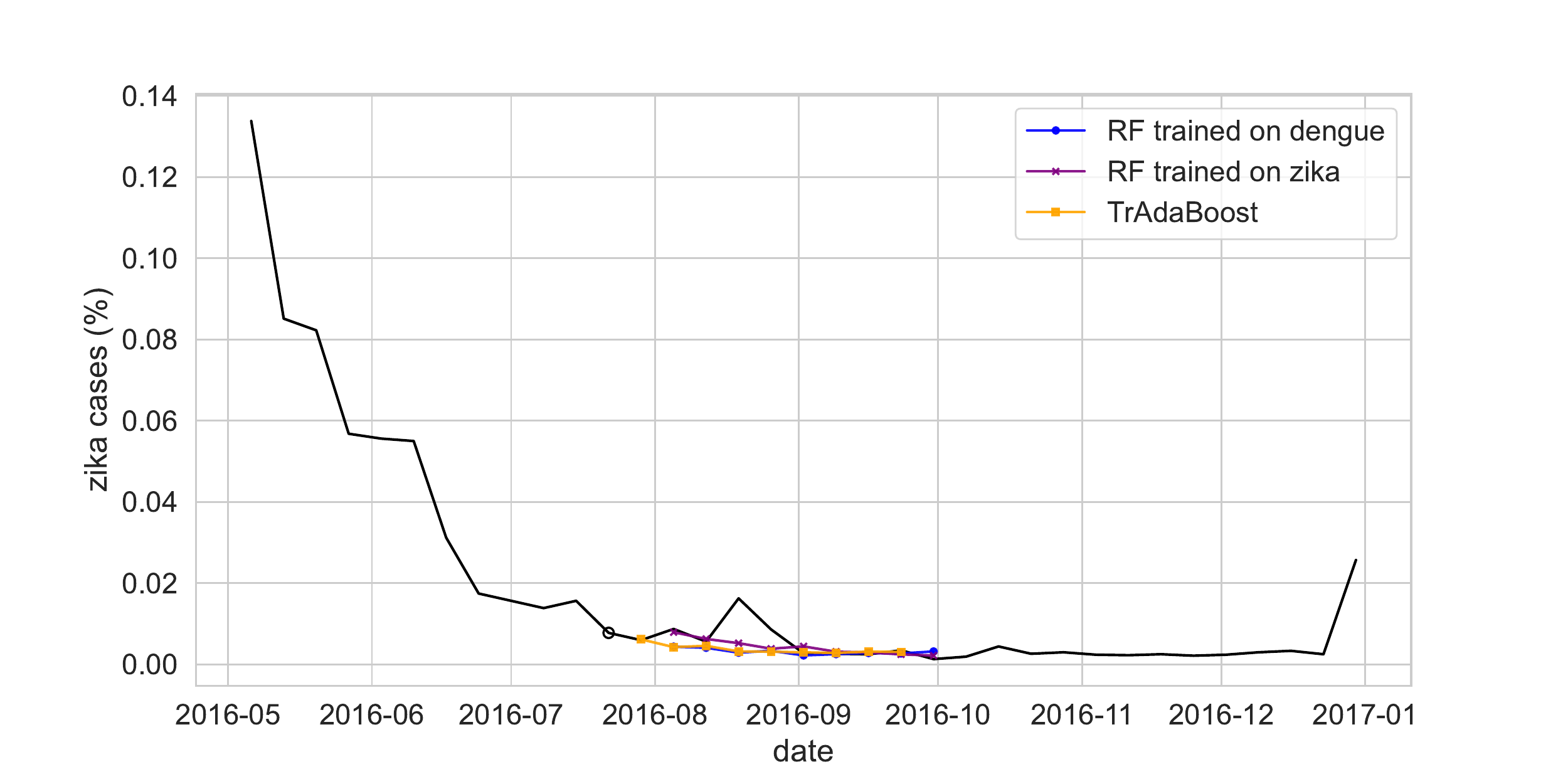}
    \caption{Random forest predictions of Zika for Rio de Janeiro at different data cutoff levels}
    \label{fig:case_study_rio_zika_rf}
\end{figure}

\begin{figure}[H]
    \centering
    \includegraphics[width=0.5\textwidth]{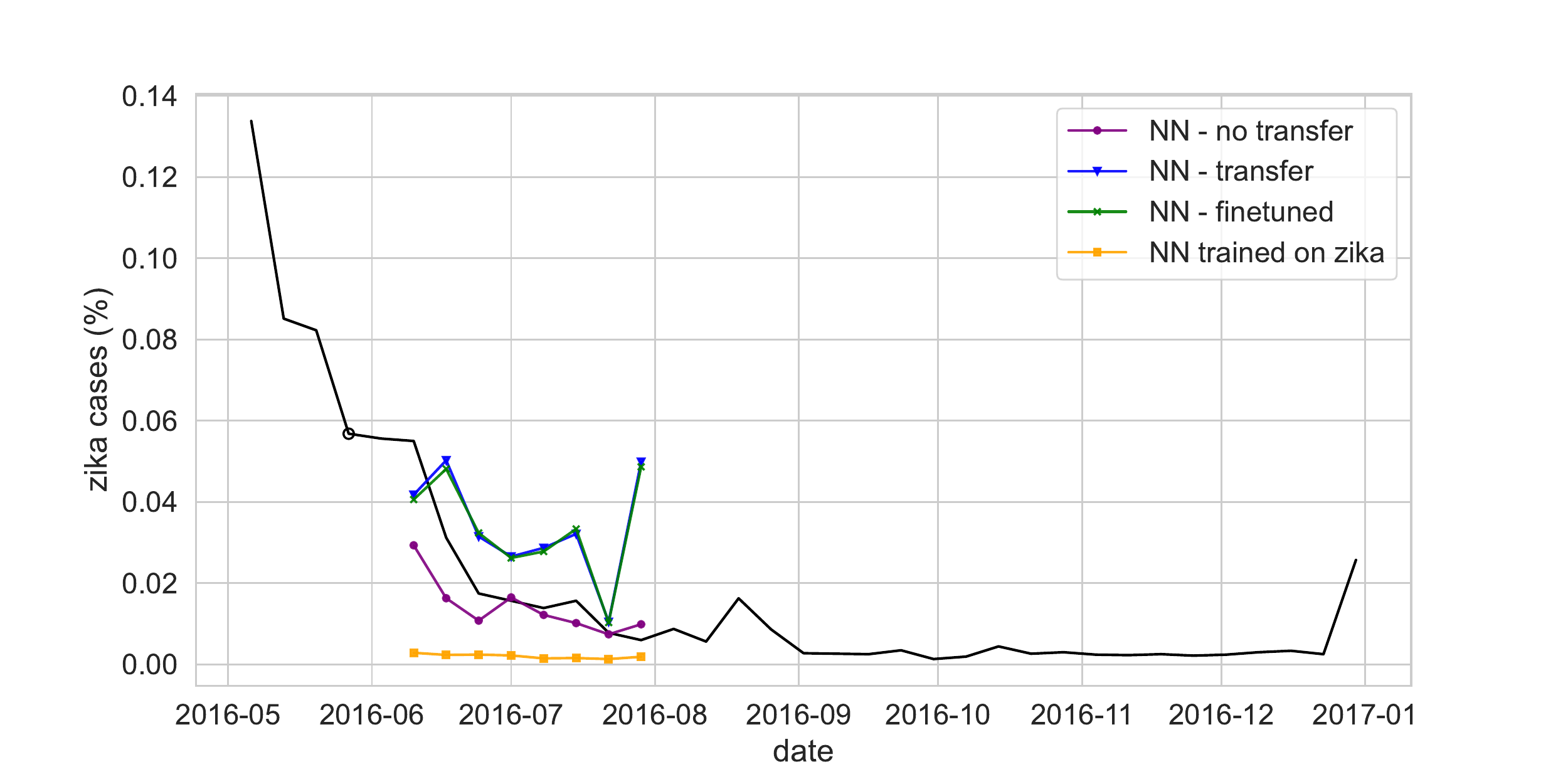}
    \includegraphics[width=0.5\textwidth]{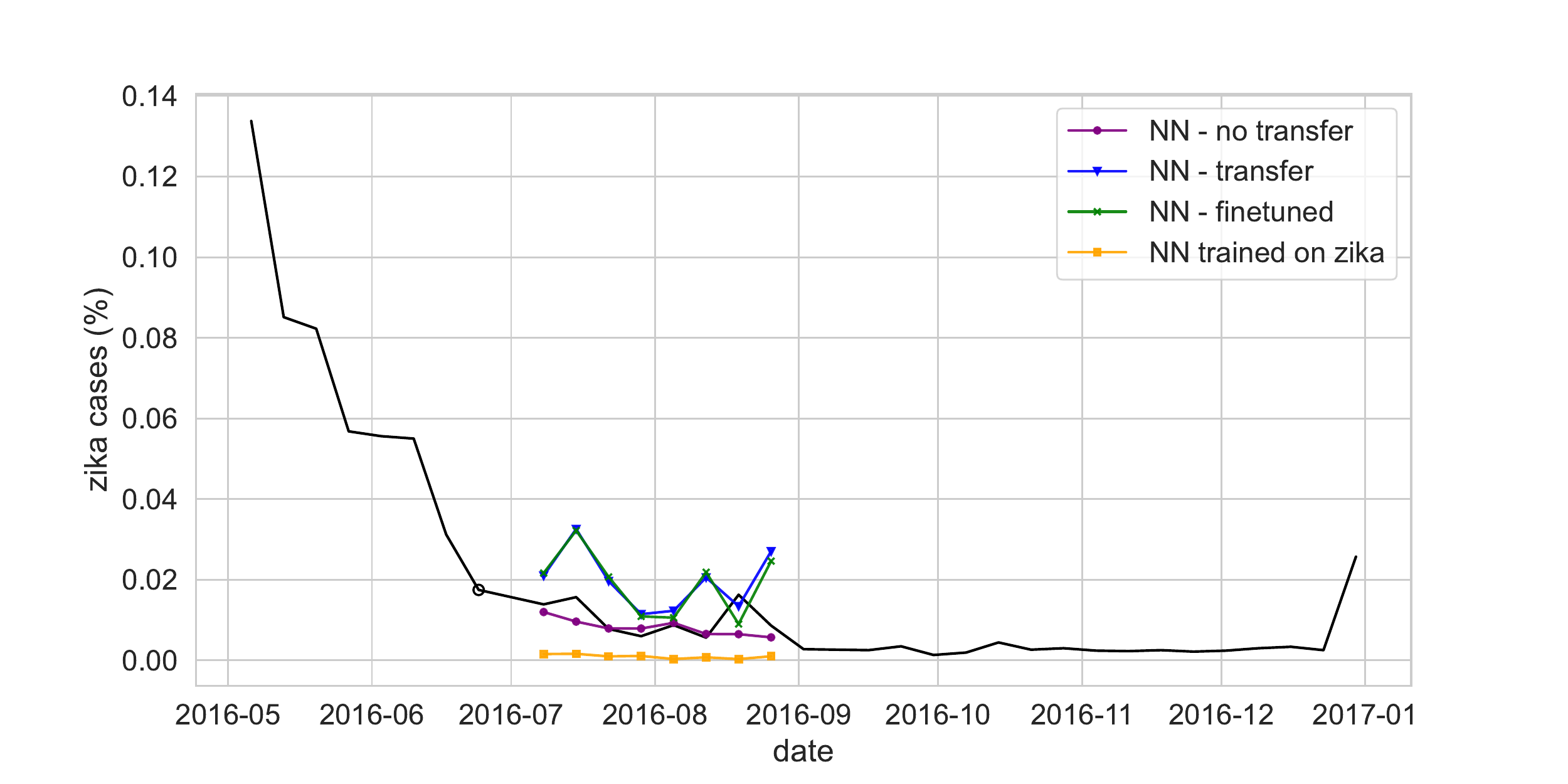}
    \includegraphics[width=0.5\textwidth]{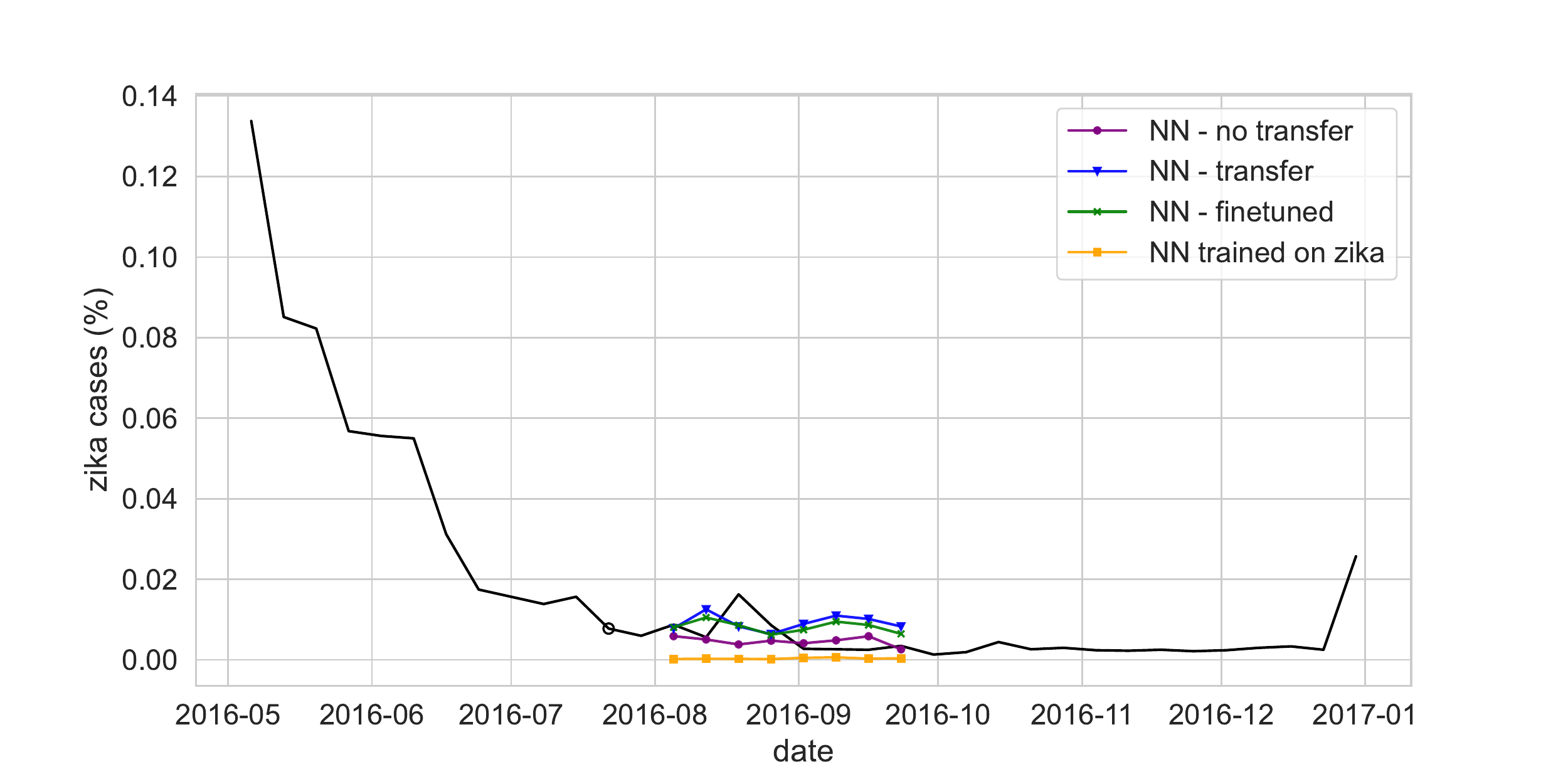}
    \caption{Neural network predictions of Zika for Rio de Janeiro}
    \label{fig:case_study_rio_zika_nn}
\end{figure}

\begin{figure}[H]
    \centering
    \includegraphics[width=0.5\textwidth]{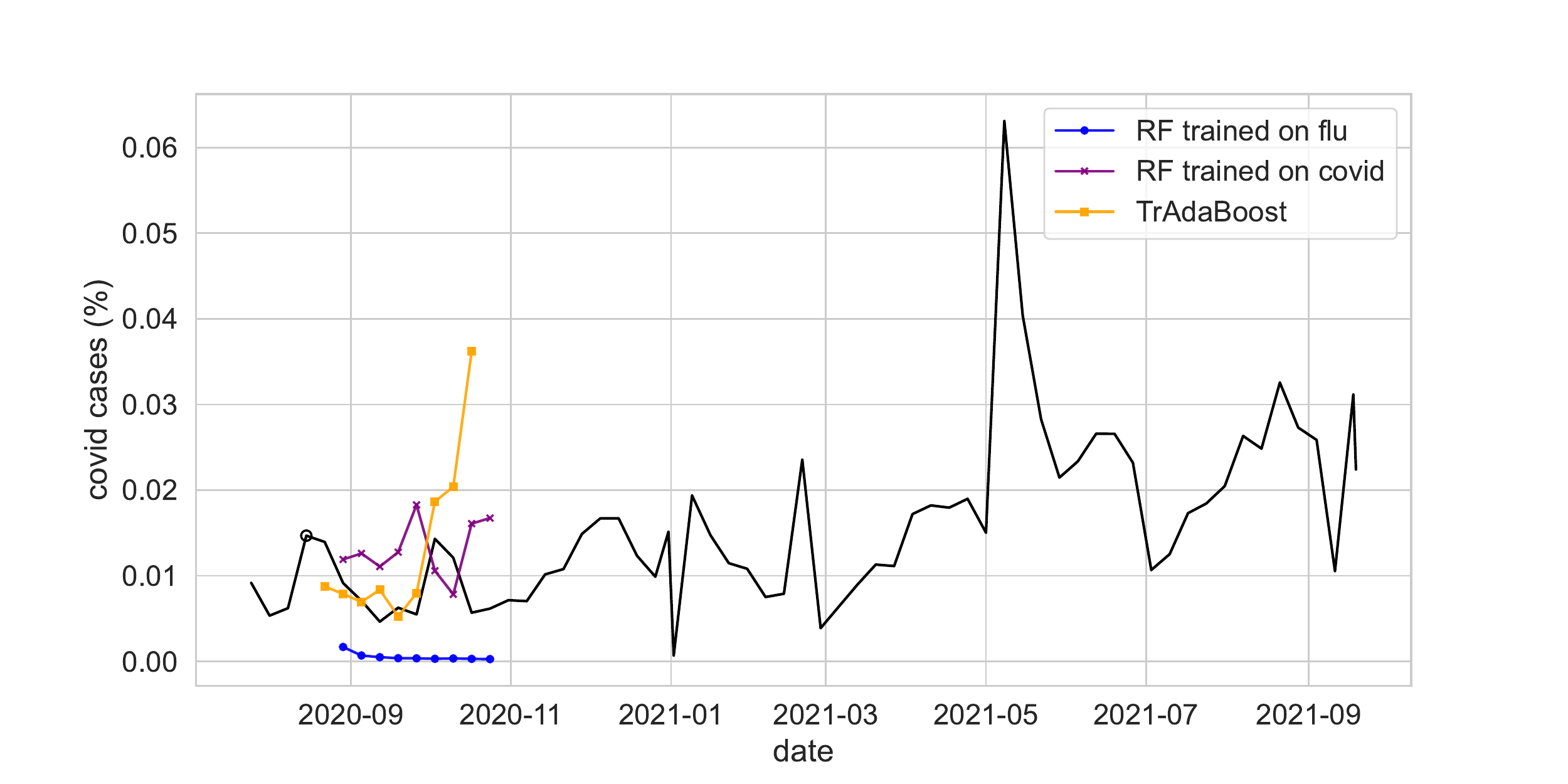}
    \includegraphics[width=0.5\textwidth]{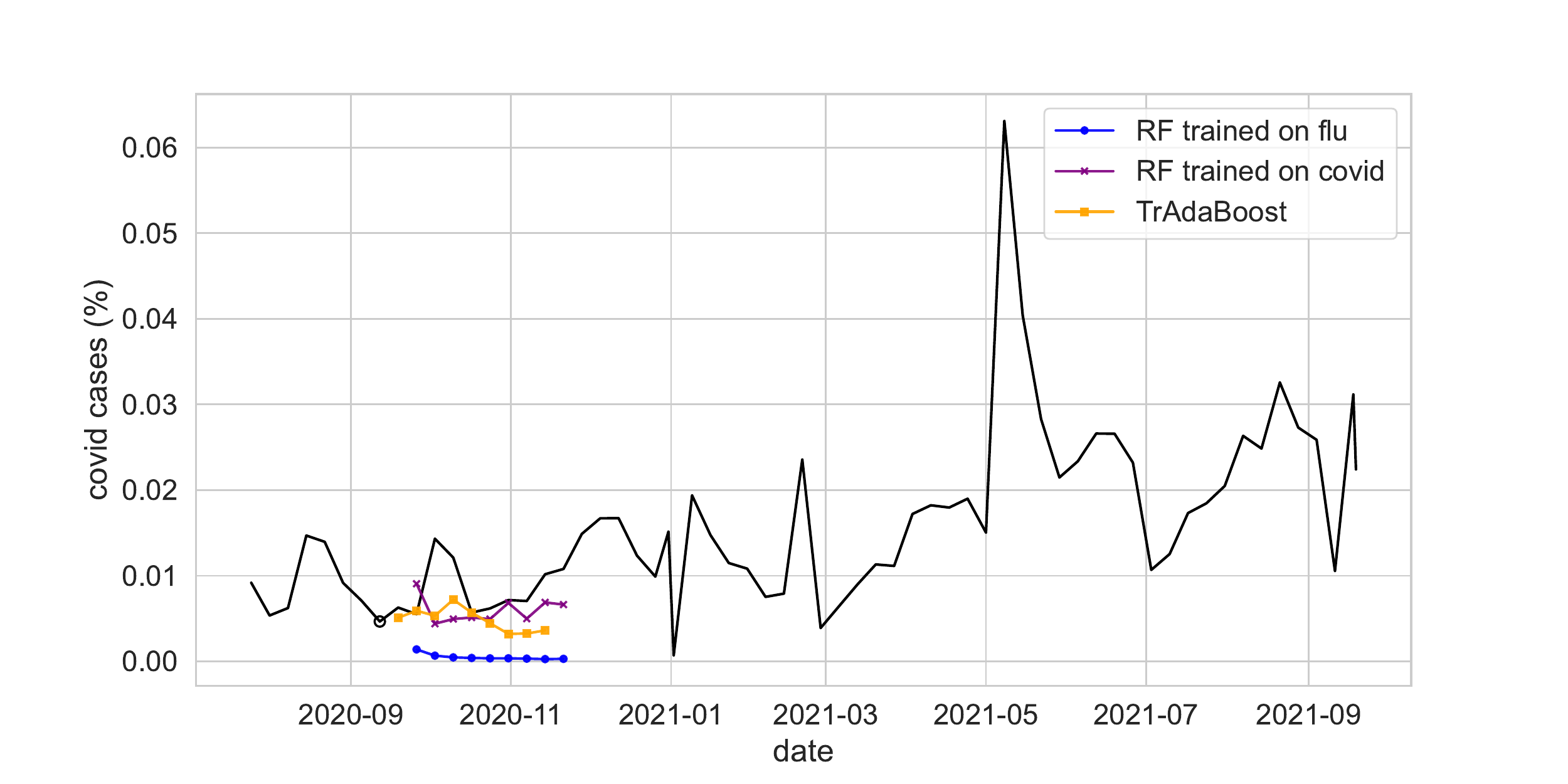}
    \includegraphics[width=0.5\textwidth]{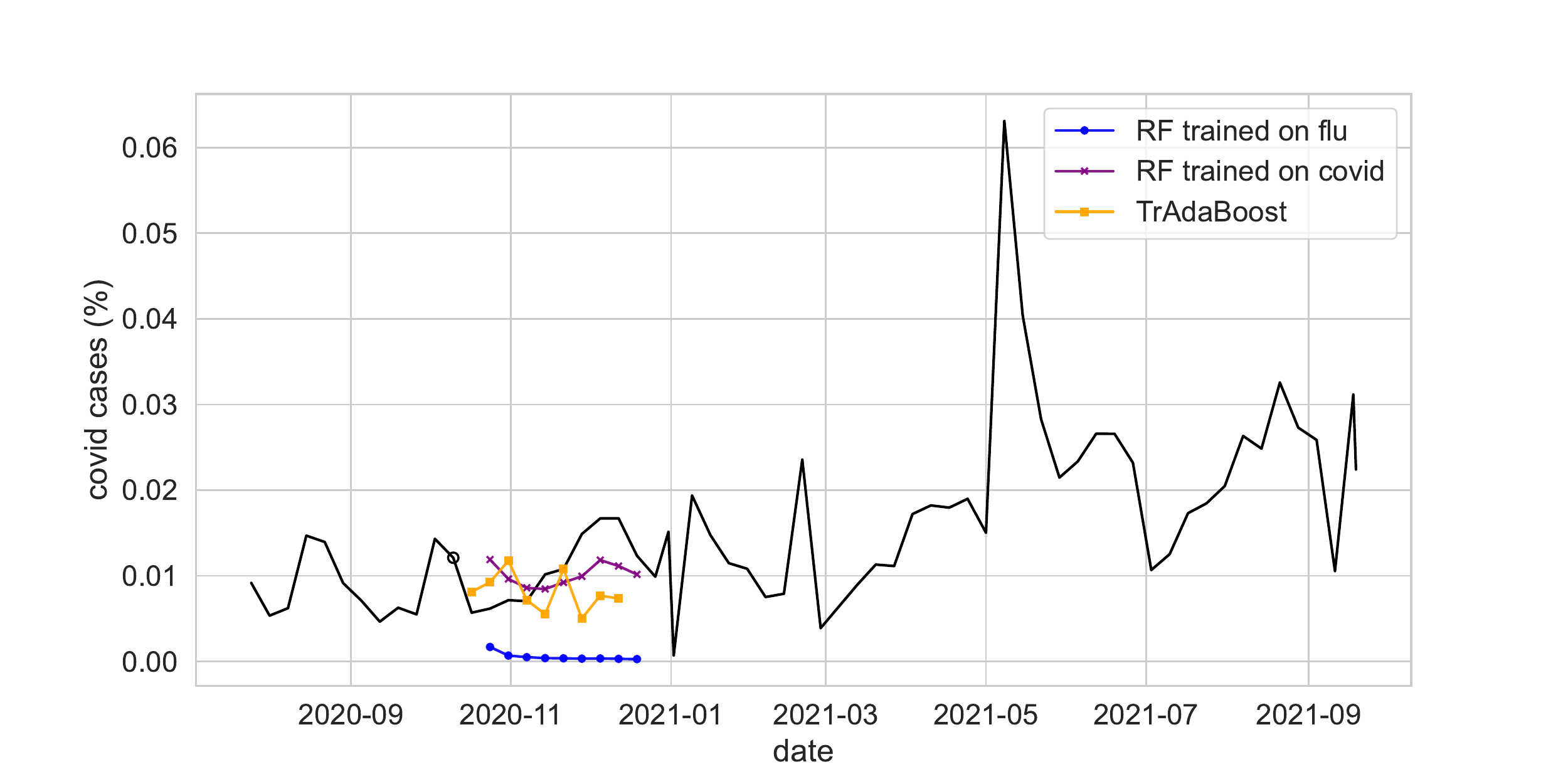}
    \caption{Random forest predictions of COVID-19 for Rio de Janeiro at different data cutoff levels}
    \label{fig:case_study_rio_covid_rf}
\end{figure}

\begin{figure}[H]
    \centering
    \includegraphics[width=0.5\textwidth]{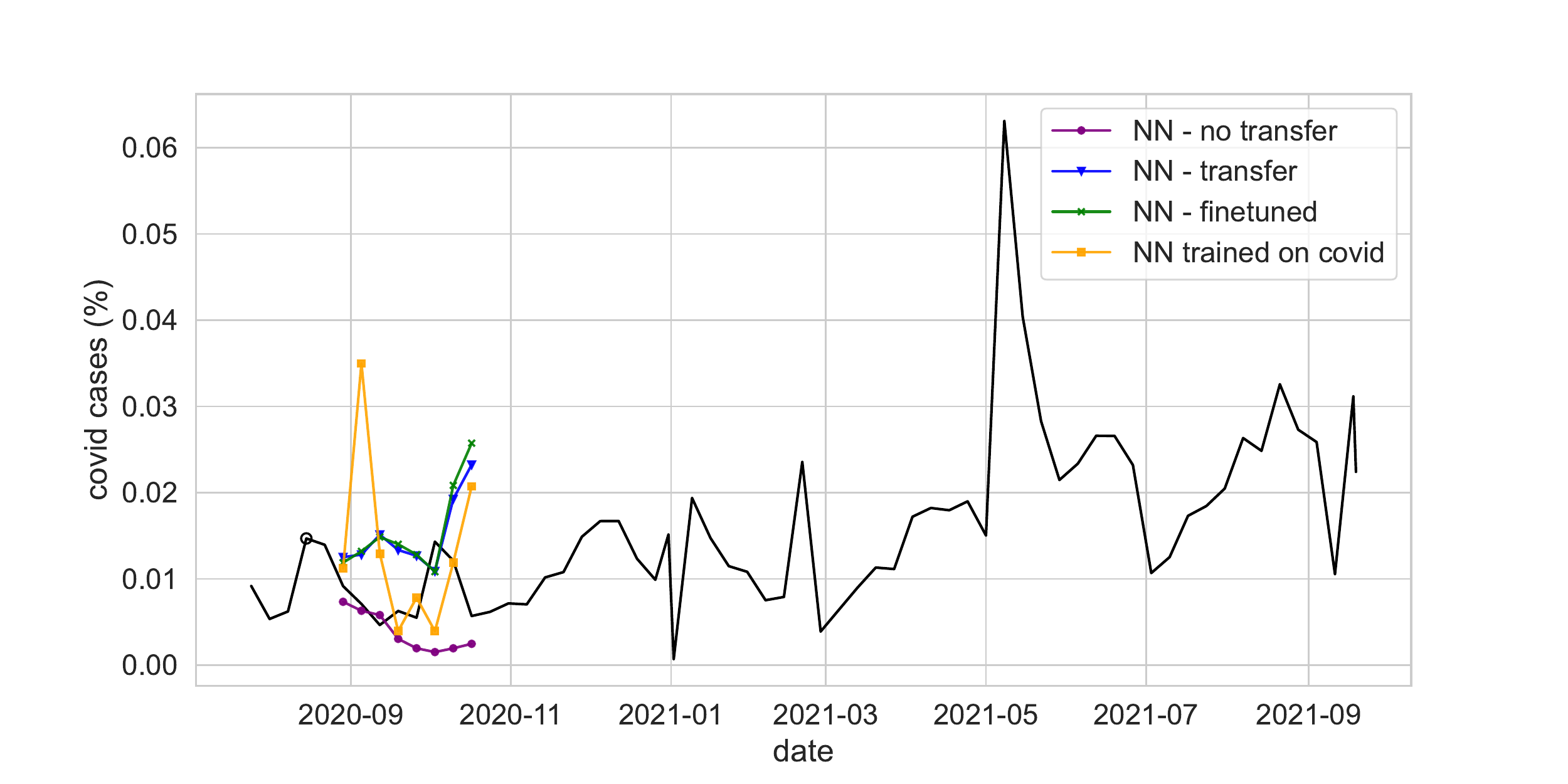}
    \includegraphics[width=0.5\textwidth]{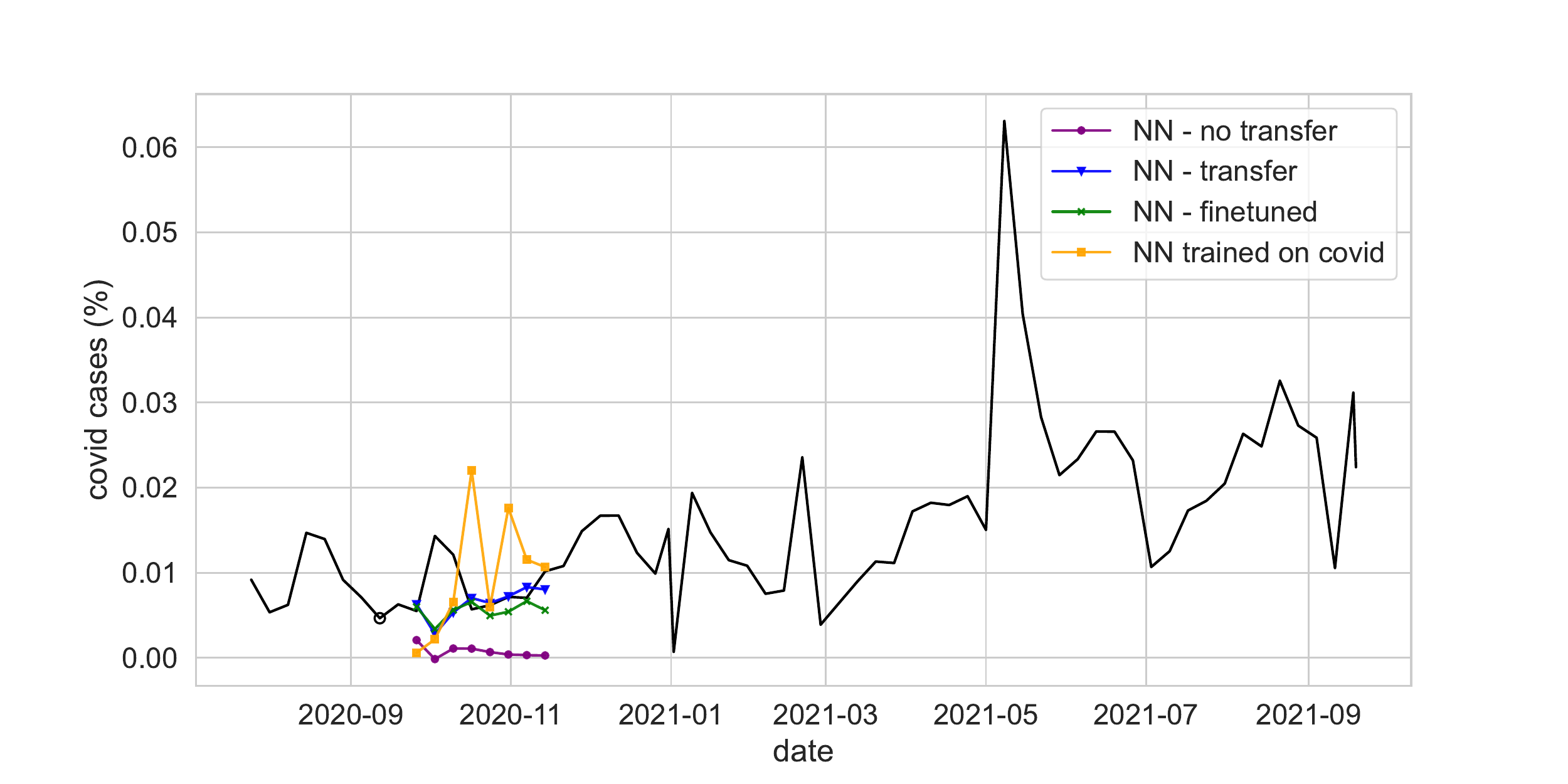}
    \includegraphics[width=0.5\textwidth]{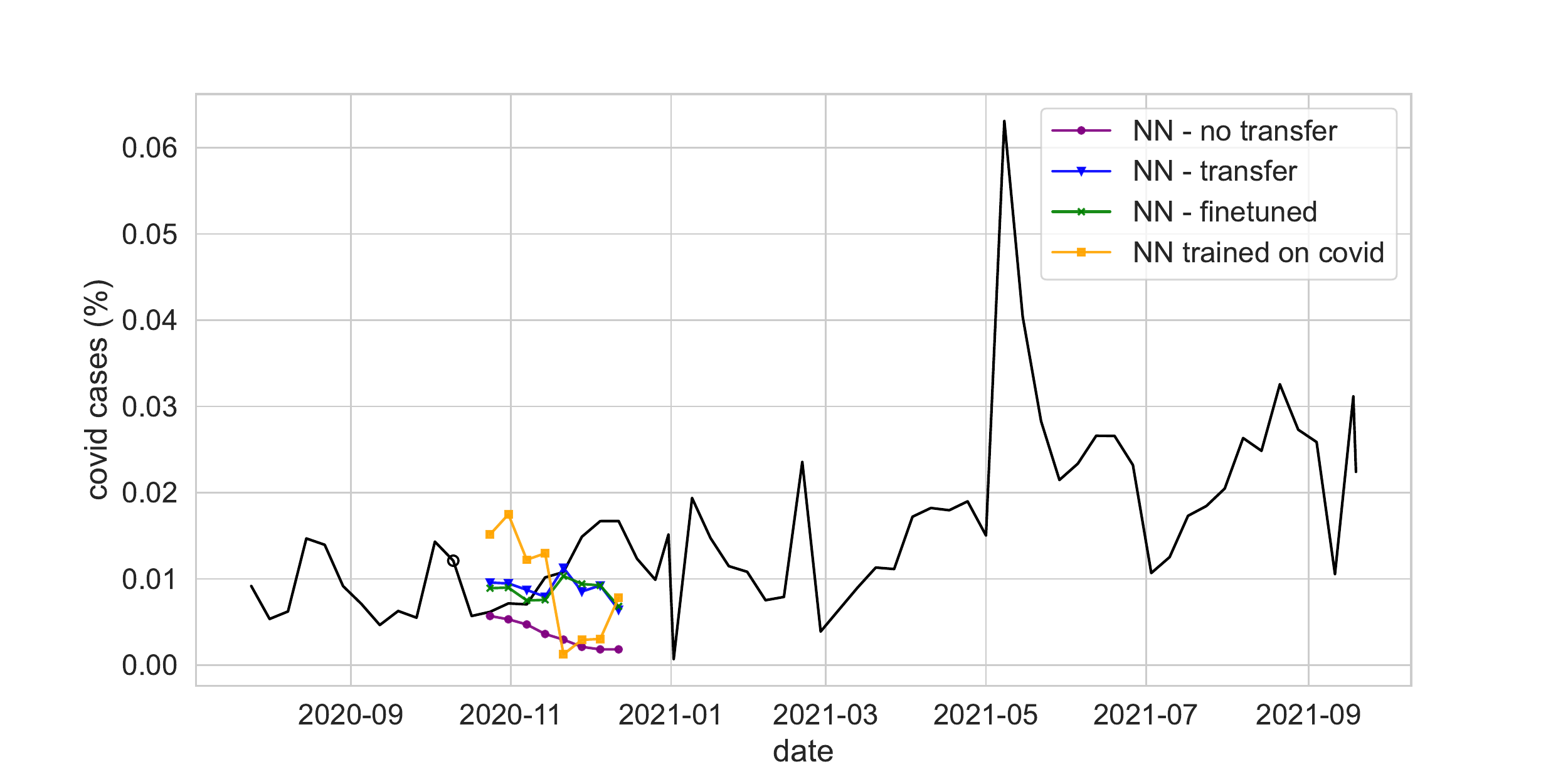}
    \caption{Neural network predictions of COVID-19 for Rio de Janeiro at different data cutoff levels}
    \label{fig:case_study_rio_covid_nn}
\end{figure}

Figures \ref{fig:case_study_rio_zika_rf}, \ref{fig:case_study_rio_zika_nn}, \ref{fig:case_study_rio_covid_rf}, and \ref{fig:case_study_rio_covid_nn} show predictions for the sample city of Rio de Janeiro.  Each plot contains one panel for each of the three data cutoff levels. The black line shows the true number of cases (expressed as percent of total cases) and the last data point used for training is highlighted as a black circle. The colored lines show the predictions for the models in each algorithm category (RF algorithms and NN algorithms). The three panels therefore simulate the predictive scenarios at different time points of the outbreak. Given training data up to the black point, e.g. May 27th 2016 in the top panel of figure \ref{fig:case_study_rio_zika_rf}, we examine which predictions the different algorithms produce and how they compare to the aspirational baselines. As the cutoffs increase (lower panels), the available data for transfer from the target disease increases, as well. Given the analysis above, we expect predictions to become more accurate as data availability increases. The figures also show how an ensemble may be composed of different transfer models, providing a prediction range for the expected number of infections.

Figure \ref{fig:case_study_rio_zika_rf} shows the strong performance of both the TrAdaBoost and the dengue-trained RF models in predicting Zika in Rio de Janeiro at all cutoff levels. The neural network transfer models are less precise (figure \ref{fig:case_study_rio_zika_nn}), though their performance approximates the dengue NN model (\textit{NN no transfer}) as the available data increases (later predictions shown in the lower panel).

In the case of COVID-19, the NN transfer and fine-tuned models produce the most consistently accurate predictions at different cutoff levels, though the RF TrAdaBoost model also performs well for cutoffs 2 and 3 (lower panels of figure \ref{fig:case_study_rio_covid_rf}). The two dengue-only models without transfer (\textit{RF} and \textit{NN - no transfer}) are less accurate.

\section{Discussion} \label{section:discussion}

We show that leveraging knowledge of related diseases via transfer learning can improve disease forecasts. This is especially relevant in the context of responding to an epidemic, where data on a new disease may be limited by resource constraints on data collection as well as the short time period since disease emergence. The transfer learning approach to epidemic forecasting may be a valuable tool not only to model emerging diseases, but also to adjust forecasts to new disease variants that may have different levels of effective contact rates or immune evasion, such as the delta or omicron variants of Sars-COV-2. It may inform policy decisions such as the scale, urgency and type of response measures.

Empirical data is useful for understanding the predictive potential of transfer learning in real-world epidemic situations. The outbreaks of both Zika and COVID-19 posed a challenge to modelers as they were unfolding.  Based on what we know about these diseases today, our study estimated whether data on dengue and influenza could have assisted in the prediction of Zika and COVID-19, respectively. The source diseases for these pairs could likely be identified by epidemiologists early on, based on transmission methods, clinical manifestations, etc. However, how an infectious disease spreads varies over time and space, with factors such as contact rates or seasonality. Empirical data offers a limited perspective and allows little extrapolation to other epidemic scenarios. Therefore, we include a theoretical analysis to compare the performance of transfer learning in different disease settings, specifically under different transmission rates, which may vary across populations or over time within the same population, and recovery rates, which also vary across populations and disease strains. 

The strong performance of the direct transfer models is noteworthy, given that models produce good predictions without incorporating any knowledge on the new disease in the training process. This finding is consistent with existing work \cite{coelho2020transfer-preprint}. Training only on the source data offers many practical benefits, including earlier deployment for disease monitoring and potentially lower computational cost. However, this study focused on diseases that are known to be fairly similar, such as the two arboviruses DEN-V and ZIK-V, which are both affected by factors like mosquito abundance. The limits of transfer without target data must be explored more broadly for diseases that are known to be dissimilar and influenced by disparate factors.

An important consideration in implementing transfer learning is the choice of machine learning algorithm and partner disease. Given the variability in best-performing models across different diseases in this study, an ensemble approach combining multiple source diseases and transfer methods may be most promising in early stages of an outbreak. The weight assigned to different models can be adjusted as more is known about the new disease. Theoretical estimates of the disease parameters such as the transmission rate as well as data-based similarity measures such as correlation can also help inform the source disease selection to achieve the best predictions. 

A limitation of this study is the difficulty of defining a baseline model. The aspirational baseline model used in this study is not ideal, since it involves extrapolation from one half of cities to the other half. An alternative approach would be to develop a baseline model trained on part of the time series for all cities and then tested on the remaining time series. However, this would restrict the test set to periods with much lower incidence than the initial outbreak peak, such as the winter in Brazil, which is the low season for Zika. This would not give an accurate estimate of the model performance in general and especially not in the context of the early stages of an epidemic. In this study, we aim to reduce the limitation of the baseline model, by randomizing the city selection across all of Brazil. This helps reduce any potential extrapolation bias.

Implementing transfer learning for epidemic response relies on some practical considerations that must be assessed on a case-by-case basis. For example:
\begin{itemize}
    \item Comparators: Which other diseases are most similar to the new disease? 
    \item Data quantity and reliability: How long has the number of cases of the new disease been measured? Is the measurement instrument likely to change and how may that affect future observation of the disease?
    \item Prediction horizon: Which prediction horizon is most useful to decision-making? How can we balance the trade-off between accuracy and early warning?
\end{itemize}

In different contexts, these questions may be easier or harder to answer. Depending on what is known about the disease, several different models can be developed to estimate a likely range of the expected number of cases.

\section{Acknowledgements}
This work was supported by grant number 2019/26595-7, São Paulo Research Foundation (FAPESP).

\bibliographystyle{elsarticle-harv}
\bibliography{references}

\begin{thebibliography}{33}
\expandafter\ifx\csname natexlab\endcsname\relax\def\natexlab#1{#1}\fi
\providecommand{\url}[1]{\texttt{#1}}
\providecommand{\href}[2]{#2}
\providecommand{\path}[1]{#1}
\providecommand{\DOIprefix}{doi:}
\providecommand{\ArXivprefix}{arXiv:}
\providecommand{\URLprefix}{URL: }
\providecommand{\Pubmedprefix}{pmid:}
\providecommand{\doi}[1]{\href{http://dx.doi.org/#1}{\path{#1}}}
\providecommand{\Pubmed}[1]{\href{pmid:#1}{\path{#1}}}
\providecommand{\bibinfo}[2]{#2}
\ifx\xfnm\relax \def\xfnm[#1]{\unskip,\space#1}\fi
\bibitem[{Altaf et~al.(2021)Altaf, Islam and Janjua}]{altaf2021novel}
\bibinfo{author}{Altaf, F.}, \bibinfo{author}{Islam, S.},
  \bibinfo{author}{Janjua, N.}, \bibinfo{year}{2021}.
\newblock \bibinfo{title}{A novel augmented deep transfer learning for
  classification of covid-19 and other thoracic diseases from x-rays}.
\newblock \bibinfo{journal}{Neural Computing and Applications}
  \bibinfo{volume}{33}.
\newblock \DOIprefix\doi{10.1007/s00521-021-06044-0}.
\bibitem[{Anastassopoulou et~al.(2020)Anastassopoulou, Russo, Tsakris and
  Siettos}]{anastassopoulou2020databased}
\bibinfo{author}{Anastassopoulou, C.}, \bibinfo{author}{Russo, L.},
  \bibinfo{author}{Tsakris, A.}, \bibinfo{author}{Siettos, C.},
  \bibinfo{year}{2020}.
\newblock \bibinfo{title}{Data-based analysis, modelling and forecasting of the
  covid-19 outbreak}.
\newblock \bibinfo{journal}{PLOS ONE} \bibinfo{volume}{15},
  \bibinfo{pages}{1--21}.
\newblock \URLprefix \url{https://doi.org/10.1371/journal.pone.0230405},
  \DOIprefix\doi{10.1371/journal.pone.0230405}.
\bibitem[{Aylett-Bullock et~al.(2020)Aylett-Bullock, Luccioni, Pham, Lam and
  Luengo-Oroz}]{bullock2020mapping}
\bibinfo{author}{Aylett-Bullock, J.}, \bibinfo{author}{Luccioni, A.},
  \bibinfo{author}{Pham, K.}, \bibinfo{author}{Lam, C.},
  \bibinfo{author}{Luengo-Oroz, M.}, \bibinfo{year}{2020}.
\newblock \bibinfo{title}{Mapping the landscape of artificial intelligence
  applications against covid-19}.
\newblock \bibinfo{journal}{JAIR} \bibinfo{volume}{69}.
\bibitem[{Biggs and Littlejohn(2021)}]{biggs2021revisiting}
\bibinfo{author}{Biggs, A.}, \bibinfo{author}{Littlejohn, L.},
  \bibinfo{year}{2021}.
\newblock \bibinfo{title}{Revisiting the initial covid-19 pandemic
  projections}.
\newblock \bibinfo{journal}{The Lancet Microbe} \bibinfo{volume}{2},
  \bibinfo{pages}{e91--e92}.
\newblock \DOIprefix\doi{10.1016/S2666-5247(21)00029-X}.
\bibitem[{Breiman(2001)}]{Breiman2001randomforests}
\bibinfo{author}{Breiman, L.}, \bibinfo{year}{2001}.
\newblock \bibinfo{title}{Random forests}.
\newblock \bibinfo{journal}{Machine Learning} \bibinfo{volume}{45},
  \bibinfo{pages}{5--32}.
\newblock \URLprefix \url{https://doi.org/10.1023/A:1010933404324},
  \DOIprefix\doi{10.1023/A:1010933404324}.
\bibitem[{CDC(2019)}]{cdc2019zika}
\bibinfo{author}{CDC}, \bibinfo{year}{2019}.
\newblock \bibinfo{title}{Zika virus, for healthcare providers, clinical
  evaluation and disease}.
\newblock \URLprefix
  \url{www.cdc.gov/zika/hc-providers/preparing-for-zika/clinicalevaluationdisease.html}.
\bibitem[{CDC(2021a)}]{cdc2021variants}
\bibinfo{author}{CDC}, \bibinfo{year}{2021}a.
\newblock \bibinfo{title}{Covid-19, what you need to know about variants}.
\newblock \URLprefix
  \url{www.cdc.gov/coronavirus/2019-ncov/variants/about-variants.html}.
\bibitem[{CDC(2021b)}]{cdc2021flaviviridae}
\bibinfo{author}{CDC}, \bibinfo{year}{2021}b.
\newblock \bibinfo{title}{Flaviviridae}.
\newblock \URLprefix \url{www.cdc.gov/vhf/virus-families/flaviviridae.html}.
\bibitem[{CDC(2021c)}]{cdc2021flu}
\bibinfo{author}{CDC}, \bibinfo{year}{2021}c.
\newblock \bibinfo{title}{Influenza (flu), information for health
  professionals}.
\newblock \URLprefix \url{www.cdc.gov/flu/professionals/index.htm}.
\bibitem[{Coelho et~al.(2020)Coelho, Holanda and
  Santos}]{coelho2020transfer-preprint}
\bibinfo{author}{Coelho, F.}, \bibinfo{author}{Holanda, N.L.},
  \bibinfo{author}{Santos, B.}, \bibinfo{year}{2020}.
\newblock \bibinfo{title}{Transfer learning applied to the forecast of
  mosquito-borne diseases}.
\newblock \bibinfo{journal}{arxiv (pre-print)}
  \DOIprefix\doi{10.1101/2020.02.03.20020164}.
\bibitem[{Desai et~al.(2019)Desai, Kraemer, Bhatia, Cori, Nouvellet, Herringer,
  Cohn, Carrion, Brownstein, Madoff and Lassmann}]{desai2019realtime}
\bibinfo{author}{Desai, A.N.}, \bibinfo{author}{Kraemer, M.U.G.},
  \bibinfo{author}{Bhatia, S.}, \bibinfo{author}{Cori, A.},
  \bibinfo{author}{Nouvellet, P.}, \bibinfo{author}{Herringer, M.},
  \bibinfo{author}{Cohn, E.L.}, \bibinfo{author}{Carrion, M.},
  \bibinfo{author}{Brownstein, J.S.}, \bibinfo{author}{Madoff, L.C.},
  \bibinfo{author}{Lassmann, B.}, \bibinfo{year}{2019}.
\newblock \bibinfo{title}{Real-time epidemic forecasting: Challenges and
  opportunities}.
\newblock \bibinfo{journal}{Health Security} \bibinfo{volume}{17},
  \bibinfo{pages}{268--275}.
\newblock \URLprefix \url{https://doi.org/10.1089/hs.2019.0022},
  \DOIprefix\doi{10.1089/hs.2019.0022}. \bibinfo{note}{pMID: 31433279}.
\bibitem[{EPA(n.d.)}]{epaIndoorAir}
\bibinfo{author}{EPA}, \bibinfo{year}{n.d.}
\newblock \bibinfo{title}{Indoor air and coronavirus (covid-19)}.
\newblock \URLprefix
  \url{www.epa.gov/coronavirus/indoor-air-and-coronavirus-covid-19}.
\bibitem[{Gautam(2021)}]{gautam2021transfer}
\bibinfo{author}{Gautam, Y.}, \bibinfo{year}{2021}.
\newblock \bibinfo{title}{Transfer learning for covid-19 cases and deaths
  forecast using lstm network}.
\newblock \bibinfo{journal}{ISA Transactions} \URLprefix
  \url{https://www.sciencedirect.com/science/article/pii/S0019057820305760},
  \DOIprefix\doi{https://doi.org/10.1016/j.isatra.2020.12.057}.
\bibitem[{Goodfellow et~al.(2016)Goodfellow, Bengio and
  Courville}]{goodfellow2016deep}
\bibinfo{author}{Goodfellow, I.}, \bibinfo{author}{Bengio, Y.},
  \bibinfo{author}{Courville, A.}, \bibinfo{year}{2016}.
\newblock \bibinfo{title}{Deep Learning}.
\newblock \bibinfo{publisher}{MIT Press}.
\newblock \bibinfo{note}{\url{http://www.deeplearningbook.org}}.
\bibitem[{Haykin(1999)}]{haykin1999neural}
\bibinfo{author}{Haykin, S.}, \bibinfo{year}{1999}.
\newblock \bibinfo{title}{Neural Networks: A Comprehensive Foundation}.
\newblock International edition, \bibinfo{publisher}{Prentice Hall}.
\newblock \URLprefix \url{https://books.google.com.br/books?id=bX4pAQAAMAAJ}.
\bibitem[{Holmdahl and Buckee(2020)}]{holmdahl2020wrong}
\bibinfo{author}{Holmdahl, I.}, \bibinfo{author}{Buckee, C.},
  \bibinfo{year}{2020}.
\newblock \bibinfo{title}{Wrong but useful — what covid-19 epidemiologic
  models can and cannot tell us}.
\newblock \bibinfo{journal}{New England Journal of Medicine}
  \bibinfo{volume}{383}, \bibinfo{pages}{303--305}.
\newblock \URLprefix \url{https://doi.org/10.1056/NEJMp2016822},
  \DOIprefix\doi{10.1056/NEJMp2016822}.
\bibitem[{Kandula et~al.(2018)Kandula, Yamana, Pei, Yang, Morita and
  Shaman}]{kandula2018evaluation}
\bibinfo{author}{Kandula, S.}, \bibinfo{author}{Yamana, T.},
  \bibinfo{author}{Pei, S.}, \bibinfo{author}{Yang, W.},
  \bibinfo{author}{Morita, H.}, \bibinfo{author}{Shaman, J.},
  \bibinfo{year}{2018}.
\newblock \bibinfo{title}{Evaluation of mechanistic and statistical methods in
  forecasting influenza-like illness}.
\newblock \bibinfo{journal}{Journal of The Royal Society Interface}
  \bibinfo{volume}{15}, \bibinfo{pages}{20180174}.
\newblock \URLprefix
  \url{https://royalsocietypublishing.org/doi/abs/10.1098/rsif.2018.0174},
  \DOIprefix\doi{10.1098/rsif.2018.0174},
  \href{http://arxiv.org/abs/https://royalsocietypublishing.org/doi/pdf/10.1098/rsif.2018.0174}{{\tt
  arXiv:https://royalsocietypublishing.org/doi/pdf/10.1098/rsif.2018.0174}}.
\bibitem[{Lipsitch and Santillana(2019)}]{Lipsitch2019EnhancingSA}
\bibinfo{author}{Lipsitch, M.}, \bibinfo{author}{Santillana, M.},
  \bibinfo{year}{2019}.
\newblock \bibinfo{title}{Enhancing situational awareness to prevent infectious
  disease outbreaks from becoming catastrophic.}
\newblock \bibinfo{journal}{Current topics in microbiology and immunology} .
\bibitem[{Maragakis(2021)}]{hopkins2021flu}
\bibinfo{author}{Maragakis, L.L.}, \bibinfo{year}{2021}.
\newblock \bibinfo{title}{Covid-19 vs. the flu}.
\newblock \URLprefix
  \url{https://www.hopkinsmedicine.org/health/conditions-and-diseases/coronavirus/coronavirus-disease-2019-vs-the-flu}.
\bibitem[{Mckibbin and Fernando(2020)}]{mckibbin2020economic}
\bibinfo{author}{Mckibbin, W.}, \bibinfo{author}{Fernando, R.},
  \bibinfo{year}{2020}.
\newblock \bibinfo{title}{The economic impact of covid-19}, in:
  \bibinfo{editor}{Richard~Baldwin, B.W.d.M.} (Ed.),
  \bibinfo{booktitle}{Economics in the Time of COVID-19}.
  \bibinfo{publisher}{Centre for Economic Policy Research}, pp.
  \bibinfo{pages}{45--51}.
\bibitem[{Pan and Yang(2010)}]{pan2010survey}
\bibinfo{author}{Pan, S.J.}, \bibinfo{author}{Yang, Q.}, \bibinfo{year}{2010}.
\newblock \bibinfo{title}{A survey on transfer learning}.
\newblock \bibinfo{journal}{IEEE Transactions on Knowledge and Data
  Engineering} \bibinfo{volume}{22}, \bibinfo{pages}{1345--1359}.
\newblock \DOIprefix\doi{10.1109/TKDE.2009.191}.
\bibitem[{Pardoe and Stone(2010)}]{pardoe2010boosting}
\bibinfo{author}{Pardoe, D.}, \bibinfo{author}{Stone, P.},
  \bibinfo{year}{2010}.
\newblock \bibinfo{title}{Boosting for regression transfer}, in:
  \bibinfo{booktitle}{Proceedings of the 27th International Conference on
  Machine Learning}, \bibinfo{publisher}{Omnipress}, \bibinfo{address}{Madison,
  WI, USA}. p. \bibinfo{pages}{863–870}.
\bibitem[{PCG et~al.(2019)PCG, RP, JC, RMR, MAP and FB}]{nunes2019thirty}
\bibinfo{author}{PCG, N.}, \bibinfo{author}{RP, D.}, \bibinfo{author}{JC,
  S.A.}, \bibinfo{author}{RMR, N.}, \bibinfo{author}{MAP, H.},
  \bibinfo{author}{FB, D.S.}, \bibinfo{year}{2019}.
\newblock \bibinfo{title}{30 years of fatal dengue cases in brazil: a review}.
\newblock \bibinfo{journal}{BMC Public Health} \bibinfo{volume}{19},
  \bibinfo{pages}{329}.
\newblock \DOIprefix\doi{10.1186/s12889-019-6641-4}.
\bibitem[{Prajapati et~al.(2017)Prajapati, Nagaraj and
  Mitra}]{prajapati2017classification}
\bibinfo{author}{Prajapati, S.A.}, \bibinfo{author}{Nagaraj, R.},
  \bibinfo{author}{Mitra, S.}, \bibinfo{year}{2017}.
\newblock \bibinfo{title}{Classification of dental diseases using cnn and
  transfer learning}, in: \bibinfo{booktitle}{2017 5th International Symposium
  on Computational and Business Intelligence (ISCBI)}, pp.
  \bibinfo{pages}{70--74}.
\newblock \DOIprefix\doi{10.1109/ISCBI.2017.8053547}.
\bibitem[{Roster et~al.(2022)Roster, Connaughton and
  Rodrigues}]{roster2021predicting}
\bibinfo{author}{Roster, K.}, \bibinfo{author}{Connaughton, C.},
  \bibinfo{author}{Rodrigues, F.A.}, \bibinfo{year}{2022}.
\newblock \bibinfo{title}{Machine learning based forecast of dengue fever in
  brazilian cities using epidemiological and meteorological variables}.
\newblock \bibinfo{journal}{American Journal of Epidemiology}
  \bibinfo{note}{(in press)}.
\bibitem[{Roy et~al.(2020)Roy, Chaudhuri, Roy, Chatterjee and
  Banerjee}]{roy2020transfer}
\bibinfo{author}{Roy, K.}, \bibinfo{author}{Chaudhuri, S.S.},
  \bibinfo{author}{Roy, P.}, \bibinfo{author}{Chatterjee, S.},
  \bibinfo{author}{Banerjee, S.}, \bibinfo{year}{2020}.
\newblock \bibinfo{title}{Transfer Learning Coupled Convolution Neural Networks
  in Detecting Retinal Diseases Using OCT Images}. \bibinfo{publisher}{Springer
  Singapore}, \bibinfo{address}{Singapore}.
\newblock pp. \bibinfo{pages}{153--173}.
\bibitem[{Saltelli et~al.(2020)Saltelli, Bammer, Bruno, Charters, Di~Fiore,
  Didier, Espeland, Kay, Lo~Piano, Mayo, Jr, Portaluri, Porter, Puy, Rafols,
  Ravetz, Reinert, Sarewitz, Stark and Vineis}]{saltelli2020five}
\bibinfo{author}{Saltelli, A.}, \bibinfo{author}{Bammer, G.},
  \bibinfo{author}{Bruno, I.}, \bibinfo{author}{Charters, E.},
  \bibinfo{author}{Di~Fiore, M.}, \bibinfo{author}{Didier, E.},
  \bibinfo{author}{Espeland, W.}, \bibinfo{author}{Kay, J.},
  \bibinfo{author}{Lo~Piano, S.}, \bibinfo{author}{Mayo, D.},
  \bibinfo{author}{Jr, R.}, \bibinfo{author}{Portaluri, T.},
  \bibinfo{author}{Porter, T.}, \bibinfo{author}{Puy, A.},
  \bibinfo{author}{Rafols, I.}, \bibinfo{author}{Ravetz, J.},
  \bibinfo{author}{Reinert, E.}, \bibinfo{author}{Sarewitz, D.},
  \bibinfo{author}{Stark, P.}, \bibinfo{author}{Vineis, P.},
  \bibinfo{year}{2020}.
\newblock \bibinfo{title}{Five ways to ensure that models serve society: a
  manifesto}.
\newblock \bibinfo{journal}{Nature} \bibinfo{volume}{582},
  \bibinfo{pages}{482--484}.
\newblock \DOIprefix\doi{10.1038/d41586-020-01812-9}.
\bibitem[{da~Saúde.(a)}]{datasus}
\bibinfo{author}{da~Saúde., M.}, a.
\newblock \bibinfo{title}{Datasus}.
\newblock \URLprefix \url{https://datasus.saude.gov.br/}.
\bibitem[{da~Saúde.(b)}]{painelcovid}
\bibinfo{author}{da~Saúde., M.}, b.
\newblock \bibinfo{title}{Painel de casos de doença pelo coronavírus 2019}.
\newblock \URLprefix \url{https://covid.saude.gov.br/}.
\bibitem[{Sufian et~al.(2020)Sufian, Ghosh, Sadiq and
  Smarandache}]{sufian2020survey}
\bibinfo{author}{Sufian, A.}, \bibinfo{author}{Ghosh, A.},
  \bibinfo{author}{Sadiq, A.S.}, \bibinfo{author}{Smarandache, F.},
  \bibinfo{year}{2020}.
\newblock \bibinfo{title}{A survey on deep transfer learning to edge computing
  for mitigating the covid-19 pandemic}.
\newblock \bibinfo{journal}{Journal of Systems Architecture}
  \bibinfo{volume}{108}, \bibinfo{pages}{101830}.
\newblock \URLprefix
  \url{https://www.sciencedirect.com/science/article/pii/S1383762120301223},
  \DOIprefix\doi{https://doi.org/10.1016/j.sysarc.2020.101830}.
\bibitem[{Tsung et~al.(2018)Tsung, Zhang, Cheng and
  Song}]{tsung2018statistical}
\bibinfo{author}{Tsung, F.}, \bibinfo{author}{Zhang, K.},
  \bibinfo{author}{Cheng, L.}, \bibinfo{author}{Song, Z.},
  \bibinfo{year}{2018}.
\newblock \bibinfo{title}{Statistical transfer learning: A review and some
  extensions to statistical process control}.
\newblock \bibinfo{journal}{Quality Engineering} \bibinfo{volume}{30},
  \bibinfo{pages}{115--128}.
\newblock \DOIprefix\doi{10.1080/08982112.2017.1373810}.
\bibitem[{WHO(2016)}]{who2016who}
\bibinfo{author}{WHO}, \bibinfo{year}{2016}.
\newblock \bibinfo{title}{Who director-general summarizes the outcome of the
  emergency committee regarding clusters of microcephaly and guillain-barré
  syndrome}.
\bibitem[{Wu et~al.(2020)Wu, Wu, Liu and Yang}]{wu2020sarscov2}
\bibinfo{author}{Wu, D.}, \bibinfo{author}{Wu, T.}, \bibinfo{author}{Liu, Q.},
  \bibinfo{author}{Yang, Z.}, \bibinfo{year}{2020}.
\newblock \bibinfo{title}{The sars-cov-2 outbreak: What we know}.
\newblock \bibinfo{journal}{International Journal of Infectious Diseases}
  \bibinfo{volume}{94}, \bibinfo{pages}{44--48}.
\newblock \URLprefix
  \url{https://www.sciencedirect.com/science/article/pii/S1201971220301235},
  \DOIprefix\doi{https://doi.org/10.1016/j.ijid.2020.03.004}.

\end{thebibliography}

\end{document}